\documentclass[pdflatex,sn-nature,iicol]{sn-jnl}

\usepackage{graphicx}%
\usepackage{multirow}%
\usepackage{amsmath,amssymb,amsfonts}%
\usepackage{amsthm}%
\usepackage{mathrsfs}%
\usepackage[title]{appendix}%
\usepackage{xcolor}%
\usepackage{textcomp}%
\usepackage{manyfoot}%
\usepackage{booktabs}%
\usepackage{algorithm}%
\usepackage{algorithmicx}%
\usepackage{algpseudocode}%
\usepackage{listings}%

\usepackage{makecell}
\usepackage{diagbox}
\usepackage{threeparttable}
\usepackage{xcolor}
\newcommand{\redbold}[1]{\color{red}\textbf{#1}}

\begin{document}

\title{Multi-Channel Masked Autoencoder and Comprehensive Evaluations for Reconstructing 12-Lead ECG from Arbitrary Single-Lead ECG}

\author[1,2,3]{\fnm{Jiarong} \sur{Chen}}\email{chenjr356@gmail.com}
\author[2]{\fnm{Wanqing} \sur{Wu}}\email{wuwanqing@mail.sysu.edu.cn}
\author[4]{\fnm{Tong} \sur{Liu}}\email{liutong@tmu.edu.cn}
\author*[1,5,6]{\fnm{Shenda} \sur{Hong}}\email{hongshenda@pku.edu.cn}
\affil*[1]{National Institute of Health Data Science, Peking University, Beijing 100191, China}
\affil[2]{School of Biomedical Engineering, Sun Yat-sen University, Shenzhen 518107, Guangdong, China}
\affil[3]{Department of Micro/Nano Electronics and MoE Key Lab of Artificial Intelligence, Shanghai Jiao Tong University, Shanghai 200240, China}

\affil[4]{Tianjin Key Laboratory of lonic-Molecular Function of Cardiovascular Disease, Department of Cardiology, Tianjin Institute of Cardiology, Second Hospital of Tianjin Medical University, Tianjin 300211, China}

\affil*[5]{Institute of Medical Technology, Health Science Center of Peking University, Beijing 100191, China }
\affil*[6]{Institute for Artificial Intelligence, Peking University, Beijing 100871, China}

\abstract{Electrocardiogram (ECG) has emerged as a widely accepted diagnostic instrument for cardiovascular diseases (CVD). The standard clinical 12-lead ECG configuration causes considerable inconvenience and discomfort, while wearable devices offers a more practical  alternative. To reduce information gap between 12-lead ECG and single-lead ECG, this study proposes a multi-channel masked autoencoder (MCMA) for reconstructing 12-Lead ECG from arbitrary single-lead ECG, and a comprehensive evaluation benchmark, ECGGenEval, encompass the signal-level, feature-level, and diagnostic-level evaluations. MCMA can achieve the state-of-the-art performance. In the signal-level evaluation, the mean square errors of 0.0317 and 0.1034, Pearson correlation coefficients of 0.7885 and 0.7420. In the feature-level evaluation, the average standard deviation of the mean heart rate across the generated 12-lead ECG is 1.0481, the coefficient of variation is 1.58\%, and the range is 3.2874. In the diagnostic-level evaluation, the average F1-score with two generated 12-lead ECG from different single-lead ECG are 0.8233 and 0.8410. }
\keywords{ECG Reconstruction, Artificial Intelligence, Deep Learning, Generated Models, Evaluation Benchmark}

\maketitle

\section{Introduction}
\label{sec:Intro}
\par The cardiovascular disease (CVD) \cite{nabel2003cardiovascular,roth2018global} contributes the leading mortality all around the world. Moreover, the prevalence rate continues to show an upward trend in the developing areas in the past decades \cite{amini2021trend}, posing a great challenge for researchers and cardiologists to address. In clinical practice, clinicians need to adopt some characterization tools \cite{holter1961new} to diagnose cardiovascular disease, and one of the most popular tools is the standard 12-lead electrocardiogram (ECG). The significant advancements in deep learning have enabled certain researchers to develop models capable of achieving cardiologist-level proficiency in interpreting 12-lead electrocardiograms (ECGs). For instance, Ribeiro et al. have successfully trained such a cardiologist-like model, as detailed in their study  \cite{ribeiro2020automatic}. In conclusion, the 12-lead ECG can provide comprehensive cardiac information from various views for doctors and classification models, playing an essential role in cardiac healthcare. 

\par However, the 12-lead ECG signal collection process puts at least 10 electrodes on the user's surface, which causes considerable inconvenience and discomfort for users, and make long-term cardiac health monitoring difficult. Up to now, the standard 12-lead ECG is traditionally used in the hospital for short-term diagnosis, usually lasting about 1 minute, while the long-term monitoring \cite{turakhia2013diagnostic} is essential  for capturing the paroxysmal cardiac abnormalities. Consequently, the pursuit of user-friendly devices capable of capturing ubiquitous electrocardiogram (ECG) signals is a priority for both researchers and markets, including patch \cite{Non_standardized,LIU2021Enhanced,turakhia2013diagnostic}, smartwatch \cite{tison2018passive,bumgarner2018smartwatch,perez2019large}, and armband \cite{rachim2016wearable,li2021Influence,lazero2020wearable}. Further, the single-lead ECG has been used for cardiac abnormality classification, such as the lead \uppercase\expandafter{\romannumeral1} ECG for the Atrial Fibrillation \cite{ENCASE}, the lead V1 ECG for the Brugada Syndrome \cite{zanchi2023identification}, and the lead aVR ECG for the Sinus Bradycardia \cite{lu2024decoding}. While wearable devices offer the advantage of ambulatory monitoring by collecting single-lead ECG signals, they do not match the diagnostic depth of a standard 12-lead ECG. The limitation arises from these devices capture the heart's electrical activity from a restricted subset of perspectives, which may not provide a comprehensive assessment of cardiac health. 

\par It is of great importance to strike a harmonious balance between clinical effectiveness and application feasibility. On the one hand, the clinical standard 12-lead ECG can comprehensively measure cardiac health \cite{ribeiro2020automatic}, but it causes somewhat inconvenience and discomfort. On the other hand, wearable devices have been a popular choice for users, but they are with limited clinical importance. Then, many researchers try to reduce the gap between the reduced-lead and 12-lead ECG, like the challenge proposed by Reyna et al. \cite{reyna2021will}. The challenge asks to access the diagnostic potential of the reduced-lead ECG, including 6-lead, 4-lead, 3-lead, and 2-lead ECG. In this challenge, Nejedly  et al. \cite{nejedly2021classification} adopt the ensemble learning, residual network, and attention mechanism to achieve state-of-the-art performance, and similarly in these researches \cite{srivastava2021channel,han2021towards,wickramasinghe2021multi,bruoth2021two}. Unfortunately, these mentioned studies only focus on the classification performance, merely providing an indirect approach to reduce the gap between the reduced-lead and 12-lead ECG.  

\par  Subsequently, some researchers try to provide a direct approach to reduce the gap between the reduce-lead (Specifically, single-lead) and 12-lead ECG, that is, reconstructing 12-lead ECG with the reduced-lead ECG \cite{edenbrandt1988vectorcardiogram,nelwan2004reconstruction,maheshwari2014accurate,atoui2010novel,sohn2020reconstruction,gundlapalle2022novel,garg2023single,lee2019synthesis,seo2022multiple,joo2023twelve}. Prior works managed to explore transformation between the Frank lead and the standard 12-lead ECG, in which the inverse Dower matrix is released by Edenbrandt et al. \cite{edenbrandt1988vectorcardiogram}, and it turns 12-lead ECG into 3-dimensional Vectorcardiogram (VCG). Nelwan et al. \cite{nelwan2004reconstruction} attempt to reconstruct 12-lead ECG from reduced lead sets. The experimental findings indicate a strong correlation coefficient of approximately 0.932 when one or two precordial leads are excluded from the lead set. Maheshwari et al. \cite{maheshwari2014accurate} adopt a solution for reconstructing 12-lead ECG from 3-lead ECG, and the reconstruction score is about 0.9187 in the testing phase. However, the assumption of dominantly linear relationship between ECG vectors can not fit the human heart electrical conduction system. Some researchers adopted autoencoders with different model architectures, such as Atoui et al. \cite{atoui2010novel} proposed Artificial Neural Network (ANN), and successfully realized the generation process of 3-lead ECG to the remaining 5 chest leads. This work and the following work all adopt the training idea of automatic encoders, including the Sohn et al. \cite{sohn2020reconstruction} used LSTM; Gundlapalle et al. \cite{gundlapalle2022novel} combined CNN and LSTM; Garg et al. \cite{garg2023single} combined the attention mechanism in autoencoder, thereby improving the feature expression ability. Generative adversarial network (GAN) \cite{goodfellow2020generative} also attracts a number of research attention, such as Lee et al. \cite{lee2019synthesis}, Seo et al. \cite{seo2022multiple} and Joo et al. \cite{joo2023twelve}. Lee et al. \cite{lee2019synthesis} adopt the conditional generative adversarial network(CGAN) to explore the feasibility of converting limb leads into chest leads. It is worth mentioned that the input of CGAN is ECG, instead of the random noise in the traditional GAN. The average structural similarity index (SSIM) between the generated ECG signal and the real ECG signal is 0.92, and the percent root mean square difference (PRD) is only 7.21\%. Seo  et al. \cite{seo2022multiple} also use the CGAN for reconstructing 12-lead, and the Mean Absolute Error (MAE) between the generated and real ECG signals is only 0.25. Joo et al. \cite{joo2023twelve} proposes a novel CGAN that consists of two generators, and achieves good reconstruction performance, like the root mean square error between the generated and real 12-lead ECG is 0.32. Additionally, our previous work \cite{zhan2024conditional} also uses this method to reconstruct 12-lead ECG from lead \uppercase\expandafter{\romannumeral1} ECG. However, the training instability and poor diversity make generating adversarial networks to difficultly address this reconstruction task, and most of the above-mentioned studies are limited flexible, since they only work on a fixed limb lead \cite{garg2023single,seo2022multiple,zhan2024conditional}. Chen et al  \cite{chen2021electrocardio} propose a novel framework to establish Electrocardio panorama; however, only the 12-lead ECG signals are considered useful, while the remaining non-standard lead signals are deemed meaningless. Consequently, there is a critical need to investigate methods for reconstructing the 12-lead ECG from an arbitrary single-lead ECG, While these methodologies are capable of approximating the reconstruction of a 12-lead electrocardiogram (ECG) from limited-lead inputs, there remains a significant research gap that needs to be addressed in the domain of 12-lead ECG reconstruction. Firstly, the traditional generative models usually focus on the fixed single-lead, instead of arbitrary single-lead ECG. Secondly, the related works \cite{atoui2010novel,sohn2020reconstruction,gundlapalle2022novel,seo2022multiple,lee2019synthesis,joo2023twelve,zhan2024conditional,garg2023single} lack a comprehensive evaluation benchmark,  mainly focus on the signal-level evaluation. Therefore, the contributions in this study are as follows:
\begin{itemize}
	\item This study proposes a multi-channel masked autoencoer, MCMA, and it can convert arbitrary single-lead ECG into the 12-lead ECG.
	\item This study designs a comprehensive evaluation benchmark, ECGGenEval, including signal-level, feature-level, and diagnostic-level evaluation.
     \item  MCMA can achieve state-of-the-art reconstruction performance in the ECGGenEval across the internal and external testing datasets, with a mean square error of 0.0317 and a Pearson correlation coefficient of 0.7885 in the internal testing dataset.
\end{itemize}

In a word, MCMA demonstrates its efficacy in reconstructing a 12-lead ECG from a single lead, thereby offering significant potential to augment the capabilities of wearable health monitoring devices in the digital health era. This advancement is poised to improve the diagnostic and monitoring capabilities of these devices, ensuring more accurate and accessible health assessments for users.

\section{Method}
\label{sec:method}
\subsection{ECG Background}
\par ECG capture the electrical activity of the heart, characterized by distinct waveforms such as the P-wave, QRS-complex, and T-wave. The standard 12-lead ECG has been a prevalent diagnostic tool in clinical practice due to its ability to provide a comprehensive view of cardiac function. This tool, however, requires the placement of 10 electrodes on the body's surface. The electrode positioning in the 12-lead ECG is detailed in Table \ref{Table:12_lead}.
\begin{table}[htbp] 
	\setlength{\tabcolsep}{1mm}{
		\caption{ECG Background: The standard electrode configuration in the standard 12-lead ECG}
		\centering
		\begin{tabular}{cc}
			\hline
			Lead & Electrode Position\\
			\hline
			\uppercase\expandafter{\romannumeral1}&Left Arm, Right Arm\\
			\uppercase\expandafter{\romannumeral2}&Left Foot, Right Arm\\
			\uppercase\expandafter{\romannumeral3}& Left Foot, Left Arm\\
			aVR&Right Arm\\
			aVL&Left Arm\\
			aVF&Left Foot\\
			V1& 4th intercostal space at the right sternal border\\
			V2& 4th intercostal space at the left sternal border\\
			V3& Midpoint between V2 and V4\\
			V4& 5th intercostal space at the midclavicular line \\
			V5&Lateral to V4, at the left midaxillary line\\
			V6&Lateral to V5, at the left midaxillary line\\
			\hline
			\label{Table:12_lead}
	\end{tabular}}
	\vspace{-0.5cm}
\end{table}
\subsection{Dataset}

\par This study conducts a large-scale 12-lead ECG datasets, consisting of 28,833 recordings from three public 12-lead ECG datasets, i.e., PTB-XL \cite{PTBXL,wagner2020ptb}, CPSC2018 \cite{CPSC2018}, and CODE-test \cite{ribeiro2020automatic}. The proposed framework is trained and validated with PTB-XL initially, and using the internal and two external testing datasets to further prove its feasibility.

\par  {PTB-XL \cite{PTBXL,wagner2020ptb} is used for model training, validating, and testing. As a large dataset, PTB-XL involves 21,799 clinical 10-second 12-lead ECG signals, and the sampling frequency is 500Hz. Based on the clinical standard, this dataset includes 71 kinds of ECG statements. As recommended, this study adopts the standard 10-fold setting, in which the folds from the 1st fold to the 8th fold is the training set, the 9th fold and the 10th fold act as the validation set and testing set, respectively. The ratio for training:validation: and testing is about 8:1:1.}

\par CPSC2018 \cite{CPSC2018} is used as an external testing set since the data distribution and information do not appear in model training and choosing. CPSC2018 contains 6,877 12-lead ECG, and these lengths varied from 6 seconds to 60 seconds with 500 Hz in sampling frequency. 

\par CODE-test is also used as an external testing set, particularly for diagnostic-level evaluation. CODE-test include 827 12-lead ECG collected from different patients with different arrhythmia. Ribeiro et al. \cite{ribeiro2020automatic} contributed a trained cardiologist-level classification model for this testing dataset.

\par Table \ref{Table:ptbxl_cpsc2018} presents the data distribution for the signal-level and feature-level evaluation in PTB-XL and CPSC2018. Table \ref{Table:code_test} presents the data distribution for the diagnostic-level evaluation in CODE-test, including 6 distinguished arrhythmia types in this dataset.

\begin{table}[htbp] 
	\setlength{\tabcolsep}{3.75mm}{
		\caption{The data distribution of PTB-XL and CPSC2018, and these datasets are used for signal-level and feature-level evaluation}
		\centering
		\begin{tabular}{ccc}
			\hline
			Dataset&Role&Number\\
			\hline
			&Training Set &87200\\
			PTB-XL&Validation Set&10965\\
			&Internal testing set&11015\\
			\hline
			CPSC2018&External testing set&55999\\
			\hline
			\label{Table:ptbxl_cpsc2018}
			
	\end{tabular}}
\end{table}

\begin{table}[htbp] 
	\setlength{\tabcolsep}{0.7mm}{
		\caption{The data distribution of CODE-test, and it is used for the diagnostic-level evaluation}
		\centering
		\begin{tabular}{cccc}
			
			\hline
			Abbreviation &Description & Quantity & \%
			\\
			\hline
			1dAVb&1st degree AV block &28&3.4\%\\
			RBBB&right bundle branch block&34&4.1\%\\
			LBBB& left bundle branch block&30&3.6\%\\
			SB& sinus bradycardia&16&1.9\%\\
			AF& atrial fibrillation &13&1.6\%\\
			ST& sinus tachycardia &36&4.4\%\\
			\hline
			\label{Table:code_test}
	\end{tabular}}
\end{table}

\subsection{MCMA}

\par
\begin{figure}[hbpt]
	\centering
	\includegraphics[width=0.45\textwidth]{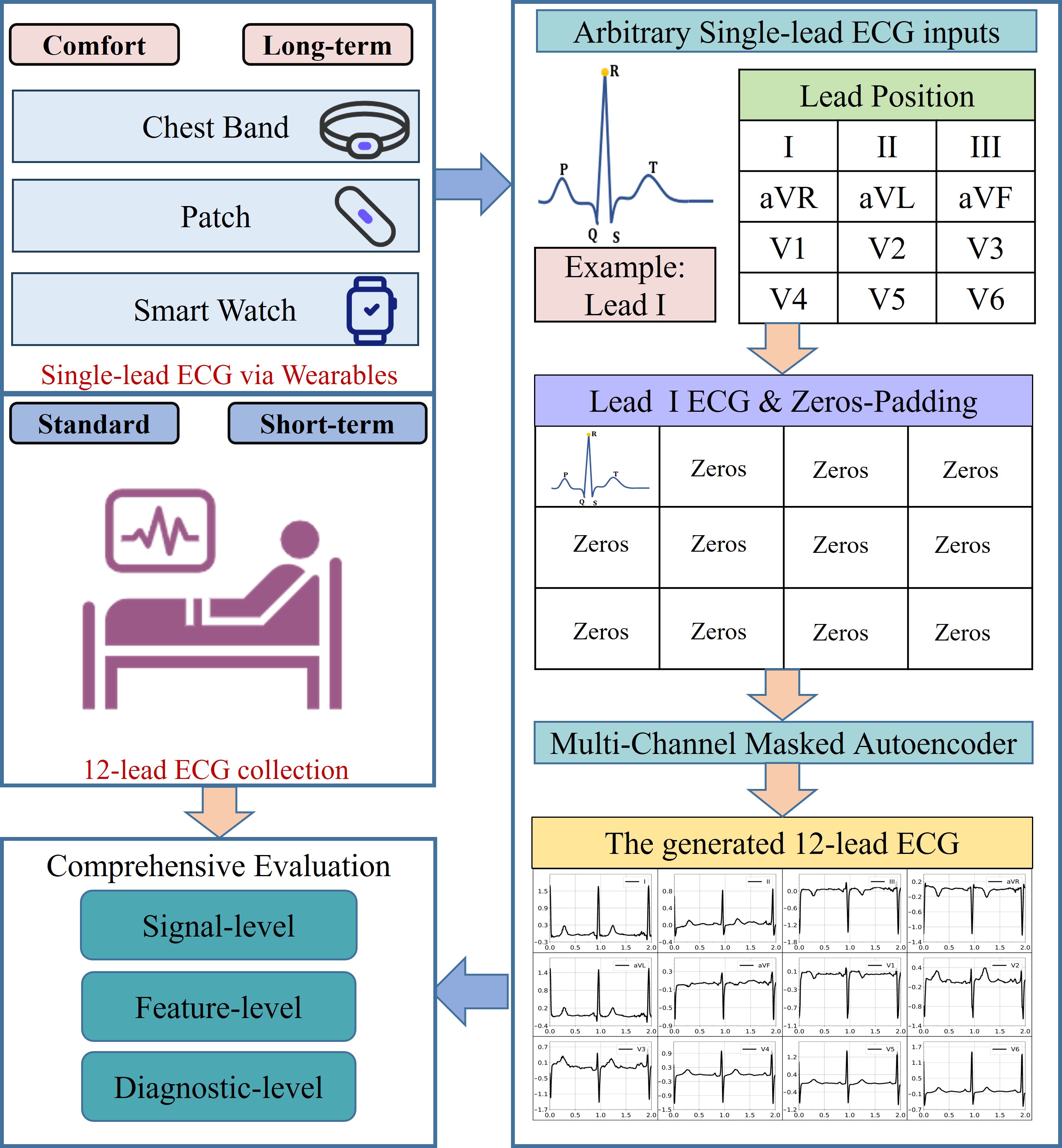}
	\caption{The 12-lead ECG generation process with single-lead ECG, the input single-lead ECG can be arbitrary, including  \uppercase\expandafter{\romannumeral1}, \uppercase\expandafter{\romannumeral2}, \uppercase\expandafter{\romannumeral3}, aVR, aVL, avF, V1, V2, V3, V4, V5, V6, and this case takes lead \uppercase\expandafter{\romannumeral1} as an example}
	\label{fig:abs}
\end{figure}

 Multi-Channel Masked Autoencoder (MCMA) masks 11 different leads, leaving only a single-lead ECG to generate the standard 12-lead ECG. MCMA takes a single-lead ECG as input and produces a 12-lead ECG as output, both with a signal length of 1024. The abstract of MCMA is seen in Fig.\ref{fig:abs}. In this study, no preprocessing steps like filtering or scaling are applied to avoid altering the ECG signals. Additionally, MCMA uses a multi-channel masked configuration to reduce training and inference costs, requiring only one model, which sets it apart from related approaches in the prior works\cite{zhan2024conditional,garg2023single,joo2023twelve,seo2022multiple}.

\subsection{Model Architecture}
\par
\begin{figure}[htbp]
	\centering
	\includegraphics[width=0.45\textwidth]{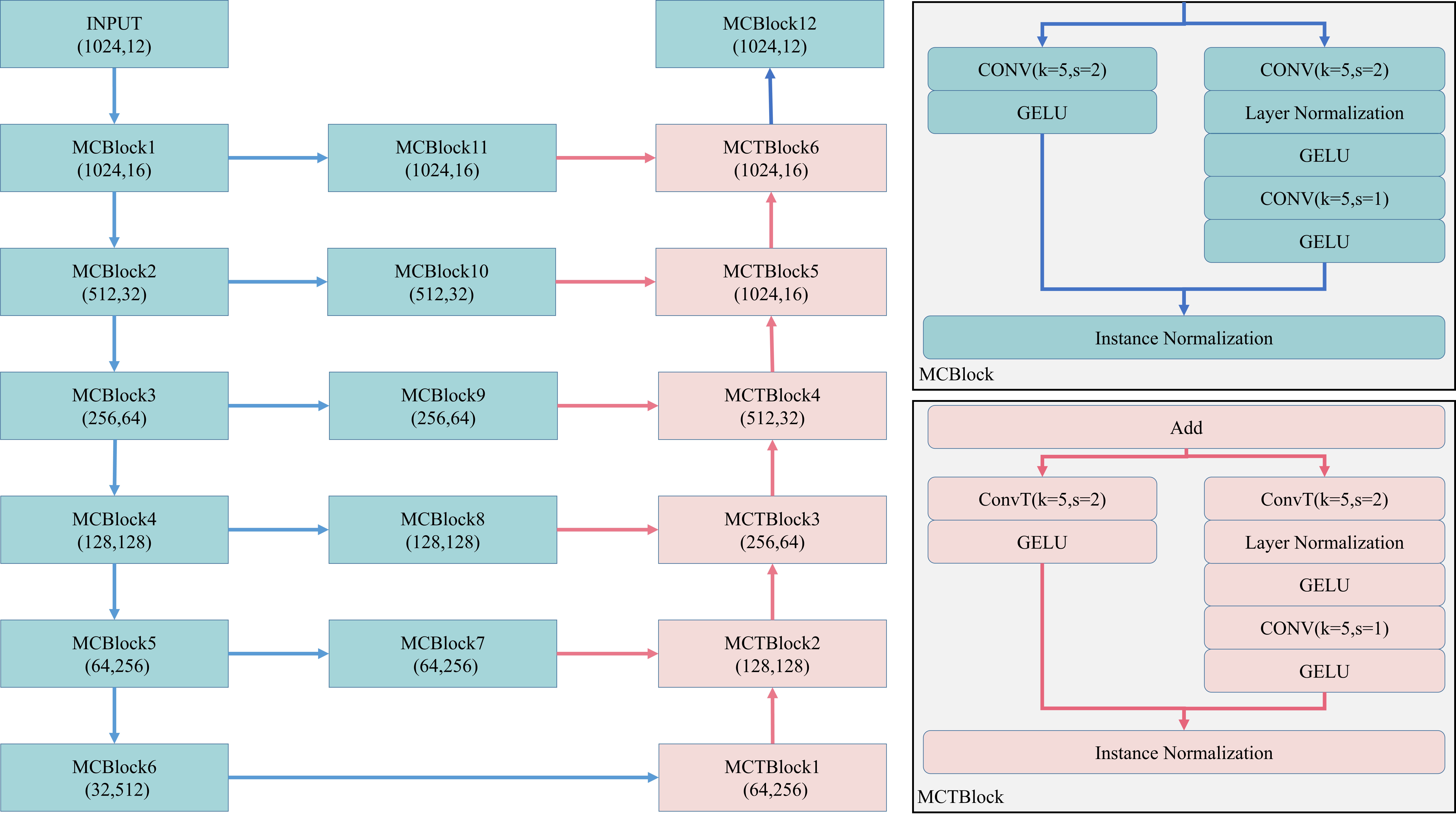}
	\caption{The detailed model architecture, the proposed  model mainly includes MCBlock and MCTBlock.}
	\label{fig:model}
\end{figure}

 MCMA needs a designed architecture, as seen in Fig.\ref{fig:model}. Motivated by ResNet \cite{43he2016deep} and UNet \cite{Unet}. The model includes two modules, namely, the downsampling and upsampling modules, which are composed of the multi-convolution block (MCBlock) and multi-convolution-transpose block (MCTBlock), respectively. The kernel size ({$k$}) is 5 and the window size ({$s$}) is 2. The choice of setting kernel size as 5 for MCBlock and MCTBlock layers aims in achieving effective feature extraction in deep learning models, particularly in those processing data with rich spatial hierarchies. The window size is usually 2 for the striding process, which can reduce the feature dimension and improve the learning ability. The activation function is GELU. The experimental results with different hyperparameters can be seen in supplementary materials. To improve the gradient stability, layer normalization (LN) and instance normalization (IN) are used in each block. The skip connections can speed up the convergence rate of the model and improve the representation ability. Additionally, the basic training recipe is provided in Table \ref{Table:hyperparamter}.

\begin{table}[htbp] 
	\setlength{\tabcolsep}{8mm}{
		\caption{The hyperparameters configuration in the MCMA training process}
		\centering
		\begin{tabular}{cc}
			\hline
			hyperparameters& configuration		
			\\
			\hline
			Batch size&256\\
			Epochs&100\\
			Signal Length&1024\\
			Optimizer&Adam\\
			Learning rate&1e-3\\
			\hline
			\label{Table:hyperparamter}
	\end{tabular}}
	\vspace{-0.5cm}
\end{table}

\subsection{MCMA Implementation}
\par \textbf{Padding Strategy:} MCMA utilize a zero-padding strategy to retain the space information for each single-lead ECG. When the single-channel ECG is processed into the 12-channel format, while the other channels are zeros, as seen in Eq.\ref{eq:zerop}.
\begin{equation}
	{P(ecg_{12},i)=I_{z}\times ecg_{12}[i]}
	\label{eq:zerop}
\end{equation}
\par In Eq.\ref{eq:zerop}, the shape of index matrix for zero-padding is {$12\times1$}, {$I_{z}(i)=1$} with other elements being zeros. Specifically, the output shape equals the input shape, and the shape of {$ecg_{12}$} is {$12\times N$}, then the shape of {$ecg_{12}[i]$} is {$1\times N$}, so the output shape also is {$12\times N$}. With zero-padding, MCMA can adaptively solve different inputs. To highlight its advantages, the 12 copies for the single-lead ECG acts as a comparison, named as the copy-padding strategy. The index matrix for copy-padding strategy, {$I_{c}$}, all elements are 1. At the same time, the arbitrary input lead and the fixed lead (lead \uppercase\expandafter{\romannumeral1}) is compared. In addition, the 12-lead ECG is provided in model training, and the padding strategy aims to mask the original 11-lead ECG with zeros or the remaining single-lead ECG in the standard 12-lead ECG. Meanwhile, only the single-lead ECG exists in the real-world application process, it should be with the padding strategy for the proposed framework.

\par\textbf{Loss Function:} The generative models mainly involve autoencoder(AE) \cite{hinton2006reducing}, generative adversarial network(GAN) \cite{goodfellow2020generative}, diffusion model \cite{DDPM}. Although the diffusion model has shown its great potential and ability in various tasks, the sampling speed \cite{DDIM} is challenging. GAN \cite{joo2023twelve,seo2022multiple,zhan2024conditional} and AE \cite{garg2023single} have been studied by the previous research works.  {Additionally, it is worth mentioning that the traditional GAN is not enough to complete this task, which supports converting random noise into the generative signals. Therefore, the researchers of this task adopted a conditional generative adversarial network, including Seo et al. \cite{seo2022multiple} Joo et el \cite{joo2023twelve}, and our previous study \cite{zhan2024conditional}.} In this study, the autoencoder can be a feasible solution for this 12-lead ECG reconstruction task, due to the training stability. Further, the proposed framework needs to compare with the GAN-based \cite{joo2023twelve,seo2022multiple,zhan2024conditional} and AE-based \cite{garg2023single} methods. 
\par The autoencoder ({$AE$}) can extract the latent representation from the raw data and convert the latent representation into the target output. The common loss function ({$L$}) is shown in Eq.\ref{eq:loss_ae}.
\begin{equation}
	L=||ecg_{12}-AE(ecg_{1})||^2 
	\label{eq:loss_ae}
\end{equation}
\par  In Eq.\ref{eq:loss_ae}, the 12-lead and single-lead ECG signals are represented by {$ecg_{12}$} and {$ecg_{1}$}. {$P$} means the padding strategy, as shown in Eq.\ref{eq:zerop}, {$i$} means the index, varying from 1 to 12. MCMA employs a zero-padding strategy as default, while copy-padding is utilized for comparative analysis within the ablation study. 
\par \textbf{Inferencing MCMA:} After the training process, MCMA can be used in real-world applications, i.e., the inferencing (testing) process. The single-lead ECG  with the zeros-padding strategy is the input of MCMA. Then, the application process for MCMA can be seen in Eq.\ref{eq:MCMA_app}.
\begin{equation}
	g_{ecg}=AE(I_{z}\times ecg_{1})
	\label{eq:MCMA_app}
\end{equation}
\par In Eq.\ref{eq:MCMA_app}, {$g_{ecg}$} is the generated 12-lead ECG with MCMA, {$ecg_{1}$} is the single-lead ECG collected by wearable devices, {$I_{z}$} can convert {$ecg_{1}$} into the input of {$AE$}.

\subsection{Comprehensive Evaluations of ECG Reconstruction}
\par This study introduces ECGGenEval, a comprehensive evaluation benchmark for 12-lead ECG reconstruction, including three distinct dimensions: signal-level, feature-level, and diagnostic-level. 
\par\textbf {Signal-Level Evaluations:} This study adopt the Pearson correlation coefficient ({$PCC$}) and mean square error ({$MSE$}) in the signal-level evaluation. The real and generated ECG signal are defined as {$r_{ecg}$} and {$g_{ecg}$}. Then, the definitions for {$PCC$} and {$MSE$} are shown in Eq.\ref{eq:PCC} and Eq.\ref{eq:MSE}.
\begin{equation}
	PCC({r_{ecg}},{g_{ecg}}) =\frac{\mu({r_{ecg}}\times{g_{ecg}} ) - \mu({r_{ecg}})\mu({g_{ecg}})}{\sigma({r_{ecg}})\sigma({g_{ecg}})}
	\label{eq:PCC}
\end{equation}
\begin{equation}
	MSE({r_{ecg}},g_{ecg})=\mu(({{r_{ecg}}-{g_{ecg}}})^{2})
	\label{eq:MSE}
\end{equation}
\par In these equations, as Eq.\ref{eq:PCC} and Eq.\ref{eq:MSE}, {$\mu(*)$} and {$\sigma(*)$} denotes the mean value and standard deviation, respectively. {$PCC$} varies from -1 to 1, and {$MSE$} is bigger than 0. The relationship between {$PCC$} and generation performance is positively related, while the relationship between {$MSE$} and generation performance is negatively related. For the signal-level evaluation, a better generative model should be with a higher {$PCC$} and lower {$MSE$}.

\par\textbf {Feature-Level Evaluations:} Furthermore, this study adopts the estimated heart rate of the generated 12-lead ECG for the feature-level evaluation. Since the heart rate in real 12-lead ECG signals theoretically occur simultaneously, and the generated signals should meet this requirement. The mean heart rate ({$MHR$}) at the {$j$}th lead can be calculated, as shown in Eq.\ref{eq:MHR}.
\begin{equation}
	MHR(j)=\frac{60\times(n-1)}{\sum_{i=1}^{n-1}{(R(i+1,j)-R(i,j))}}
	\label{eq:MHR}
\end{equation}
\par In Eq.\ref{eq:MHR}, the {$i$}th detected R-wave in {$j$}th lead is denoted as {$R(i,j)$}, and its unit is second. Therefore, {$MHR$} can represent the heartbeat per minute. Based on the 12 {$MHR$} from different 12-lead ECG, the average value {$MMHR$} can be computed with Eq.\ref{eq:mmhr}. Then, the feature-level evaluation involves standard deviation ({$SD$}), Range (the difference between maximum and minimum), and coefficient of variation ({$CV$}), expressed as {$MHR_{SD}$}, {$MHR_{Range}$} and {$MHR_{CV}$}, respectively. The calculation processes can be seen in Eq.\ref{eq:mhr_sd}, Eq.\ref{eq:mhr_range} and Eq.\ref{eq:mhr_cv}, respectively.
\begin{equation}
	{MMHR}=\frac{1}{12} \sum_{j=1}^{12} (MHR(j)
	\label{eq:mmhr}
\end{equation}
\begin{equation}
	MHR_{SD}=\sqrt{\frac{1}{12} \sum_{j=1}^{12} (MHR(j) - MMHR)^2}
	\label{eq:mhr_sd}
\end{equation}
\begin{equation}
	MHR_{Range}=\max(MHR)-\min(MHR)
	\label{eq:mhr_range}
\end{equation}
\begin{equation}
	MHR_{CV}=\frac{MHR_{SD}}{MMHR}
	\label{eq:mhr_cv}
\end{equation}
\par The reference estimation is completed with the original 12-lead ECG These feature-level evaluation is good if the inter-lead heart rates are consistent.
\par \textbf {Diagnostic-Level Evaluations:} This study also adopts the diagnostic-level evaluation for this 12-lead ECG reconstruction task. MCMA is able to convert the limit-lead (even single-lead) ECG into 12-lead ECG, which bridges the limited-lead ECG to the classifiers which trained with 12-lead ECG as input. Therefore, this study can evaluate the  generated 12-lead ECG using classification performance, including the precision ({$Pre$}), recall ({$Rec$}), specificity ({$Spe$}) and F1-score ({$F_{1}$}), as shown in literature \cite{ribeiro2020automatic}. These calculation process of classification metric are seen in Eq.\ref{eq:Pre}, Eq.\ref{eq:Rec}, Eq.\ref{eq:Spe} and Eq.\ref{eq:F1}. 
\begin{equation}
	{Pre}=\frac{TP}{TP+FP}
	\label{eq:Pre}
\end{equation}
\begin{equation}
	{Rec}=\frac{TP}{TP+FN}
	\label{eq:Rec}
\end{equation}
\begin{equation}
	{Spe}=\frac{TN}{TN+FP}
	\label{eq:Spe}
\end{equation}
\begin{equation}
	{F_{1}}=\frac{2\times TP}{2\times TP+FN+FP}
	\label{eq:F1}
\end{equation}
\par Also, the original classification performance with the real 12-lead ECG is the standard reference, and the generated 12-lead ECG with the other methods\cite{garg2023single,zhan2024conditional,seo2022multiple,joo2023twelve} are used in the result comparison.

\section{Results}

\subsection{Signal-Level Performance}


\par First of all, the signal-level evaluation is the primary evaluation metric, such as {$MSE$} and {$PCC$}. In contrast to conventional approaches, this scheme offers a distinct advantage: it enables the conversion of an arbitrary single-lead ECG to a 12-lead ECG without the necessity of training multiple generative models. The experimental results of {$MSE$} and {$PCC$} are shown in Table \ref{Table:ptbxl_mse_pcc}, where the horizontal direction represents the output and the vertical direction represents the input. Besides, the reconstruction performance in the external dataset, CPSC2018, is seen in Table \ref{Table:cpsc_mse_pcc}.

\begin{table*}[t!]
	\centering
	\caption{The signal-level evaluation of mean square error ({$MSE$}) and Pearson correlation coefficient ($PCC$) between the generated and real 12-lead ECG in the internal testing set, PTB-XL}
	\setlength{\tabcolsep}{1.5mm}{
		\tiny
		\begin{tabular}{c|ccccccccccccc}
			\Xhline{1.5pt}
			
			\diagbox{Input}{Output}& \uppercase\expandafter{\romannumeral1}& \uppercase\expandafter{\romannumeral2}& \uppercase\expandafter{\romannumeral3}&aVR&aVL&aVF&V1&V2&V3&V4&V5&V6&Mean	
			\\
			\hline
			\hline
			&\multicolumn{13}{c}{{$MSE$}}\\
			\hline
			\uppercase\expandafter{\romannumeral1}& 0.0071 & 0.0166 & 0.0219 & 0.0064 & 0.0106 & 0.0175 & 0.0326 & 0.0679 & 0.0698 & 0.0482 & 0.0398 & 0.0491 & 0.0323  \\ \hline
			\uppercase\expandafter{\romannumeral2} & 0.0119 & 0.0093 & 0.0203 & 0.0058 & 0.0138 & 0.0118 & 0.0341 & 0.0691 & 0.0721 & 0.0480 & 0.0388 & 0.0483 & 0.0319  \\ \hline
			\uppercase\expandafter{\romannumeral3} & 0.0113 & 0.0138 & 0.0156 & 0.0090 & 0.0098 & 0.0115 & 0.0345 & 0.0703 & 0.0766 & 0.0569 & 0.0457 & 0.0521 & 0.0339  \\ \hline
			aVR & 0.0095 & 0.0117 & 0.0255 & 0.0046 & 0.0143 & 0.0160 & 0.0325 & 0.0670 & 0.0686 & 0.0457 & 0.0364 & 0.0469 & 0.0316  \\ \hline
			aVL & 0.0087 & 0.0160 & 0.0169 & 0.0084 & 0.0088 & 0.0139 & 0.0338 & 0.0685 & 0.0732 & 0.0534 & 0.0436 & 0.0511 & 0.0330  \\ \hline
			aVF & 0.0123 & 0.0108 & 0.0167 & 0.0077 & 0.0117 & 0.0106 & 0.0345 & 0.0700 & 0.0749 & 0.0532 & 0.0429 & 0.0505 & 0.0330  \\ \hline
			V1 & 0.0117 & 0.0172 & 0.0268 & 0.0082 & 0.0147 & 0.0188 & 0.0251 & 0.0592 & 0.0689 & 0.0543 & 0.0438 & 0.0505 & 0.0332  \\ \hline
			V2 & 0.0125 & 0.0173 & 0.0255 & 0.0088 & 0.0146 & 0.0182 & 0.0307 & 0.0431 & 0.0546 & 0.0494 & 0.0450 & 0.0518 & 0.0309  \\ \hline
			V3 & 0.0120 & 0.0171 & 0.0261 & 0.0084 & 0.0147 & 0.0183 & 0.0325 & 0.0548 & 0.0410 & 0.0395 & 0.0407 & 0.0507 & 0.0296  \\ \hline
			V4 & 0.0115 & 0.0160 & 0.0254 & 0.0078 & 0.0145 & 0.0175 & 0.0335 & 0.0630 & 0.0548 & 0.0283 & 0.0345 & 0.0485 & 0.0296  \\ \hline
			V5 & 0.0112 & 0.0146 & 0.0253 & 0.0070 & 0.0146 & 0.0168 & 0.0334 & 0.0670 & 0.0644 & 0.0377 & 0.0267 & 0.0457 & 0.0304  \\ \hline
			V6 & 0.0112 & 0.0143 & 0.0254 & 0.0069 & 0.0146 & 0.0166 & 0.0330 & 0.0683 & 0.0698 & 0.0434 & 0.0329 & 0.0393 & 0.0313  \\ \hline
			Mean & 0.0109 & 0.0145 & 0.0226 & 0.0074 & 0.0130 & 0.0156 & 0.0325 & 0.0640 & 0.0657 & 0.0465 & 0.0392 & 0.0487 & 0.0317  \\ \hline
			\hline
			\hline
			&\multicolumn{13}{c}{$PCC$}\\
			\hline
			
			\uppercase\expandafter{\romannumeral1}  & 0.9876 & 0.7588 & 0.5714 & 0.9167 & 0.8374 & 0.5516 & 0.8297 & 0.7950 & 0.7646 & 0.8075 & 0.8430 & 0.8358 & 0.7916  \\ \hline
			\uppercase\expandafter{\romannumeral2}   & 0.8358 & 0.9897 & 0.6738 & 0.9366 & 0.6519 & 0.8887 & 0.8095 & 0.7833 & 0.7516 & 0.8167 & 0.8599 & 0.8572 & 0.8212  \\ \hline
			\uppercase\expandafter{\romannumeral3}   & 0.8428 & 0.8283 & 0.9824 & 0.8168 & 0.8999 & 0.9048 & 0.7969 & 0.7747 & 0.7130 & 0.7383 & 0.7803 & 0.7879 & 0.8222  \\ \hline
			aVR & 0.9090 & 0.9055 & 0.3998 & 0.9890 & 0.6237 & 0.6356 & 0.8331 & 0.7970 & 0.7756 & 0.8324 & 0.8810 & 0.8759 & 0.7881  \\ \hline
			aVL & 0.9262 & 0.7650 & 0.8838 & 0.8407 & 0.9815 & 0.7475 & 0.8106 & 0.7884 & 0.7367 & 0.7635 & 0.8009 & 0.8025 & 0.8206  \\ \hline
			aVF & 0.8193 & 0.9317 & 0.8951 & 0.8637 & 0.7729 & 0.9848 & 0.7987 & 0.7769 & 0.7257 & 0.7696 & 0.8118 & 0.8162 & 0.8305  \\ \hline
			V1 & 0.8353 & 0.7357 & 0.3209 & 0.8474 & 0.5896 & 0.4856 & 0.9800 & 0.8508 & 0.7601 & 0.7557 & 0.7996 & 0.8110 & 0.7310  \\ \hline
			V2 & 0.8210 & 0.7326 & 0.3891 & 0.8268 & 0.6114 & 0.5174 & 0.8695 & 0.9898 & 0.8760 & 0.7994 & 0.7941 & 0.7947 & 0.7518  \\ \hline
			V3 & 0.8362 & 0.7416 & 0.3565 & 0.8423 & 0.6029 & 0.5141 & 0.8375 & 0.8947 & 0.9886 & 0.8868 & 0.8404 & 0.8154 & 0.7631  \\ \hline
			V4 & 0.8495 & 0.7771 & 0.4038 & 0.8659 & 0.6097 & 0.5687 & 0.8172 & 0.8302 & 0.8792 & 0.9887 & 0.9077 & 0.8544 & 0.7793  \\ \hline
			V5 & 0.8603 & 0.8157 & 0.4045 & 0.8944 & 0.6088 & 0.5983 & 0.8184 & 0.7992 & 0.8065 & 0.9065 & 0.9864 & 0.9059 & 0.7838  \\ \hline
			V6 & 0.8596 & 0.8284 & 0.3976 & 0.8992 & 0.6077 & 0.6104 & 0.8242 & 0.7892 & 0.7685 & 0.8568 & 0.9240 & 0.9861 & 0.7793  \\ \hline
			Mean & 0.8652 & 0.8175 & 0.5566 & 0.8783 & 0.6998 & 0.6673 & 0.8354 & 0.8224 & 0.7955 & 0.8268 & 0.8524 & 0.8452 & 0.7885  \\ \hline
			\Xhline{1.5pt}
	\end{tabular}}
	\label{Table:ptbxl_mse_pcc}
\end{table*}
\begin{table*}[h]
	\centering
	\caption{ The signal-level evaluation of mean square error ({$MSE$}) and Pearson correlation coefficient ({$PCC$}) between the generated and real 12-lead ECG in an external testing set, CPSC2018}
	
	\setlength{\tabcolsep}{1.5mm}{
		\tiny
		\begin{tabular}{c|ccccccccccccc}
			\Xhline{1.5pt}
			\diagbox{Input}{Output}& \uppercase\expandafter{\romannumeral1}& \uppercase\expandafter{\romannumeral2}& \uppercase\expandafter{\romannumeral3}&aVR&aVL&aVF&V1&V2&V3&V4&V5&V6&Mean
			\\
			\hline
			\hline
			&\multicolumn{13}{c}{{$MSE$}}\\
			\hline
			\uppercase\expandafter{\romannumeral1} & 0.0311 & 0.0424 & 0.0447 & 0.0321 & 0.0322 & 0.0406 & 0.0892 & 0.1277 & 0.1544 & 0.1707 & 0.2133 & 0.2700 & 0.1040  \\ \hline
			\uppercase\expandafter{\romannumeral2}  & 0.0369 & 0.0339 & 0.0413 & 0.0312 & 0.0364 & 0.0336 & 0.0938 & 0.1340 & 0.1592 & 0.1716 & 0.2135 & 0.2698 & 0.1046  \\ \hline
			\uppercase\expandafter{\romannumeral3}  & 0.0369 & 0.0386 & 0.0357 & 0.0345 & 0.0315 & 0.0335 & 0.0938 & 0.1362 & 0.1662 & 0.1823 & 0.2229 & 0.2755 & 0.1073  \\ \hline
			aVR & 0.0353 & 0.0376 & 0.0474 & 0.0280 & 0.0375 & 0.0381 & 0.0889 & 0.1309 & 0.1558 & 0.1691 & 0.2119 & 0.2693 & 0.1041  \\ \hline
			aVL & 0.0337 & 0.0423 & 0.0380 & 0.0349 & 0.0295 & 0.0368 & 0.0914 & 0.1314 & 0.1625 & 0.1818 & 0.2230 & 0.2754 & 0.1067  \\ \hline
			aVF & 0.0381 & 0.0354 & 0.0371 & 0.0333 & 0.034 & 0.0324 & 0.0967 & 0.1376 & 0.1648 & 0.1792 & 0.2195 & 0.2735 & 0.1068  \\ \hline
			V1 & 0.0377 & 0.0436 & 0.0489 & 0.0332 & 0.0376 & 0.0415 & 0.0743 & 0.1172 & 0.1549 & 0.1801 & 0.2223 & 0.2747 & 0.1055  \\ \hline
			V2 & 0.0377 & 0.0428 & 0.0483 & 0.0343 & 0.0375 & 0.0408 & 0.0853 & 0.0903 & 0.1328 & 0.1708 & 0.2213 & 0.2750 & 0.1014  \\ \hline
			V3 & 0.0372 & 0.0422 & 0.0483 & 0.0340 & 0.0376 & 0.0405 & 0.0883 & 0.1063 & 0.1139 & 0.1565 & 0.2140 & 0.2724 & 0.0993  \\ \hline
			V4 & 0.0368 & 0.0413 & 0.0485 & 0.0331 & 0.0376 & 0.0401 & 0.0907 & 0.1187 & 0.1318 & 0.1387 & 0.2051 & 0.2687 & 0.0993  \\ \hline
			V5 & 0.0363 & 0.0407 & 0.0479 & 0.0326 & 0.0373 & 0.0397 & 0.0912 & 0.1269 & 0.1454 & 0.1555 & 0.1867 & 0.2643 & 0.1004  \\ \hline
			V6 & 0.0362 & 0.0406 & 0.0478 & 0.0324 & 0.0373 & 0.0397 & 0.0914 & 0.1303 & 0.1534 & 0.1636 & 0.2041 & 0.2461 & 0.1019  \\ \hline
			Mean & 0.0362 & 0.0401 & 0.0445 & 0.0328 & 0.0355 & 0.0381 & 0.0896 & 0.1240 & 0.1496 & 0.1683 & 0.2131 & 0.2696 & 0.1034  \\ \hline
			\hline
			&\multicolumn{13}{c}{{$PCC$}}\\
			\hline
			\uppercase\expandafter{\romannumeral1}  & 0.9823 & 0.7645 & 0.4415 & 0.8997 & 0.7368 & 0.5368 & 0.7334 & 0.7064 & 0.7289 & 0.8097 & 0.8521 & 0.8515 & 0.7536  \\ \hline
			\uppercase\expandafter{\romannumeral2} & 0.7857 & 0.9874 & 0.5924 & 0.9347 & 0.4796 & 0.8900 & 0.6677 & 0.6464 & 0.6868 & 0.8044 & 0.8505 & 0.8521 & 0.7648  \\ \hline
			\uppercase\expandafter{\romannumeral3} & 0.7816 & 0.8380 & 0.9766 & 0.8098 & 0.8553 & 0.8997 & 0.6642 & 0.6321 & 0.6312 & 0.7212 & 0.7655 & 0.7752 & 0.7792  \\ \hline
			aVR & 0.8758 & 0.9172 & 0.3065 & 0.9867 & 0.4574 & 0.6816 & 0.7219 & 0.6822 & 0.7265 & 0.8353 & 0.8807 & 0.8846 & 0.7464  \\ \hline
			aVL & 0.8671 & 0.7468 & 0.8291 & 0.7985 & 0.9718 & 0.7056 & 0.7005 & 0.6745 & 0.6604 & 0.7262 & 0.7685 & 0.7783 & 0.7689  \\ \hline
			aVF & 0.7477 & 0.9331 & 0.8683 & 0.8546 & 0.6577 & 0.9815 & 0.6240 & 0.6115 & 0.6393 & 0.7459 & 0.7934 & 0.7999 & 0.7714  \\ \hline
			V1 & 0.7797 & 0.7248 & 0.2136 & 0.8242 & 0.4177 & 0.4808 & 0.9701 & 0.7794 & 0.7140 & 0.7381 & 0.7751 & 0.7916 & 0.6841  \\ \hline
			V2 & 0.7638 & 0.7313 & 0.2186 & 0.8143 & 0.4311 & 0.4913 & 0.7959 & 0.9842 & 0.8725 & 0.8020 & 0.7893 & 0.7897 & 0.7070  \\ \hline
			V3 & 0.7844 & 0.7565 & 0.2331 & 0.8350 & 0.4279 & 0.5279 & 0.7453 & 0.8749 & 0.9851 & 0.9022 & 0.8503 & 0.8277 & 0.7292  \\ \hline
			V4 & 0.7981 & 0.7881 & 0.2372 & 0.8609 & 0.4241 & 0.5602 & 0.7051 & 0.7763 & 0.8883 & 0.9871 & 0.9156 & 0.8698 & 0.7342  \\ \hline
			V5 & 0.8191 & 0.8073 & 0.2755 & 0.8847 & 0.4394 & 0.5825 & 0.7047 & 0.7172 & 0.7954 & 0.9143 & 0.9863 & 0.9167 & 0.7369  \\ \hline
			V6 & 0.8233 & 0.8098 & 0.2809 & 0.8906 & 0.4390 & 0.5867 & 0.7048 & 0.6883 & 0.7405 & 0.8630 & 0.9254 & 0.9866 & 0.7282  \\ \hline
			Mean & 0.8174 & 0.8171 & 0.4561 & 0.8661 & 0.5615 & 0.6604 & 0.7281 & 0.7311 & 0.7557 & 0.8208 & 0.8461 & 0.8436 & 0.7420  \\ \hline
			\Xhline{1.5pt}
	\end{tabular}}
	\label{Table:cpsc_mse_pcc}
	\vspace{1.45mm}
\end{table*}
\subsection{Feature-Level Performance}
\par This study also provides the feature-level evaluation results for MCMA, including the standard deviation {$MHR_{SD}$}, Range {$MHR_{Range}$} and coefficient of variation {$MHR_{CV}$}. The feature-level evaluation results in the internal testing set PTB-XL and external testing set CPSC2018 are shown in Table \ref{Table:ptbxl_mhr} and Table \ref{Table:cpsc_mhr}, respectively. In the mentioned two tables, the first group is the reference value of the original 12-lead ECG. Additionally, the R-peak recognition is completed by algorithm \cite{Makowski2021neurokit}.

\begin{table}[htbp]
	\centering
	\caption{The feature-level evaluation for the generated and real 12-lead ECG in the internal testing dataset, PTBXL}
	
	\setlength{\tabcolsep}{2mm}{
		
		\begin{tabular}{c|ccc}
			\Xhline{1.5pt}
			\diagbox{Input}{Metric}&{$MHR_{SD}$}&{$MHR_{CV}$}&{$MHR_{Range}$}\\
			\hline
			Original&2.2137&3.21\%&7.2195\\\hline
			\uppercase\expandafter{\romannumeral1} & 1.1164 & 1.63\% & 3.5415  \\ 
			\uppercase\expandafter{\romannumeral2} & 1.0073 & 1.51\% & 3.0958  \\ 
			\uppercase\expandafter{\romannumeral3} & 1.2042 & 1.80\% & 3.7227  \\ 
			aVR & 1.0785 & 1.61\% & 3.3192  \\ 
			aVL & 1.1893 & 1.77\% & 3.6891  \\ 
			aVF & 0.934 & 1.40\% & 2.8122  \\ 
			V1 & 0.9944 & 1.51\% & 3.1111  \\ 
			V2 & 0.9582 & 1.48\% & 3.0753  \\ 
			V3 & 1.0777 & 1.66\% & 3.4607  \\ 
			V4 & 0.9287 & 1.43\% & 2.9728  \\ 
			V5 & 1.0264 & 1.56\% & 3.2796  \\ 
			V6 & 1.0627 & 1.61\% & 3.3682  \\ \hline
			Mean&1.0481&1.58\%&3.2874\\
			\Xhline{1.5pt}
	\end{tabular}}
	\label{Table:ptbxl_mhr}
\end{table}

\begin{table}[htbp]
	\centering
	\caption{ {The feature-level evaluation for the generated and real 12-lead ECG in the external testing dataset, CPSC2018}}
	
	\setlength{\tabcolsep}{2mm}{
		
		\begin{tabular}{c|ccc}
			\Xhline{1.5pt}
			\diagbox{Input}{Metric}&{$MHR_{SD}$}&{$MHR_{CV}$}&{$MHR_{Range}$}\\
			\hline
			Original&2.1313&2.65\%&7.1267\\ \hline
			\uppercase\expandafter{\romannumeral1} & 1.3510 & 1.70\% & 4.4984  \\ 
			\uppercase\expandafter{\romannumeral2} & 0.7732 & 0.99\% & 2.4776  \\ 
			\uppercase\expandafter{\romannumeral3} & 0.9133 & 1.17\% & 2.9432  \\ 
			aVR & 0.8590 & 1.10\% & 2.7977  \\ 
			aVL & 1.3069 & 1.69\% & 4.2717  \\ 
			aVF & 0.8146 & 1.04\% & 2.5931  \\ 
			V1 & 0.9355 & 1.24\% & 3.0275  \\ 
			V2 & 0.8629 & 1.16\% & 2.8255  \\ 
			V3 & 0.8467 & 1.16\% & 2.7630  \\ 
			V4 & 0.8529 & 1.14\% & 2.7957  \\ 
			V5 & 0.9544 & 1.28\% & 3.1345  \\ 
			V6 & 0.9093 & 1.22\% & 2.9712  \\ \hline
			Mean&0.9483&1.24\%&3.0916\\
			\Xhline{1.5pt}
	\end{tabular}}
	\label{Table:cpsc_mhr}
\end{table}

\subsection{Diagnostic-Level Performance}
\par Lastly, this study demonstrate the diagnostic-level performance of MCMA. The classifier is trained and validated by Ribeiro  et al. \cite{ribeiro2020automatic}, which only accepts the 12-lead ECG. Then, it is essential to present the classification performance with the generated 12-lead ECG. For example, Table \ref{Table:classification_12_1} shows the classification performance of the generated 12-lead ECG with lead \uppercase\expandafter{\romannumeral1}. The detailed diagnostic-level evaluations are shown in Table \ref{Table:mean_diagnostic}, including the original 12-lead ECG (as the reference), the single-lead ECG (i.e., MCMA input) and the generated 12-lead ECG (i.e., MCMA output), which directly shows the gain in the arrhythmia classification task.

\begin{table}[hbp] 
	\setlength{\tabcolsep}{1.8mm}{
		
		\caption{ {The diagnostic-level evaluation for MCMA, as the generated 12-lead ECG is from lead \uppercase\expandafter{\romannumeral1} ECG, CODE-test}}
		\centering
		\begin{tabular}{ccccc}
			
			\Xhline{1.5pt}
			\diagbox{Class}{Metric}&{$Pre$}&{$Rec$}&{$Spe$}&{$F_{1}$}\\
			\hline
			1dAVb & 0.8750 & 0.7500 & 0.9962 & 0.8077  \\ \hline
			RBBB & 0.8788 & 0.8529 & 0.9950 & 0.8657  \\ \hline
			LBBB & 0.9630 & 0.8667 & 0.9987 & 0.9123  \\ \hline
			SB & 0.7273 & 1.0000 & 0.9926 & 0.8421  \\ \hline
			AF & 0.5833 & 0.5385 & 0.9939 & 0.5600  \\ \hline
			ST & 0.9459 & 0.9459 & 0.9975 & 0.9459  \\ \hline
			Mean & 0.8289 & 0.8257 & 0.9956 & 0.8223  \\ \hline
			\Xhline{1.5pt}
			\label{Table:classification_12_1}
	\end{tabular}}
	
\end{table}

\begin{table}[htbp]
	\centering
	\caption{ {The diagnostic-level evaluation for the generated 12-lead ECG in another external testing dataset, CODE-test }}
	
	\setlength{\tabcolsep}{1mm}{
		
		\begin{tabular}{c|cccc}
			\Xhline{1.5pt}
			\diagbox{Input}{Metric}&{$Pre$}&{$Rec$}&{$Spe$}&{$F_{1}$}\\
			\hline
			Original 12-lead \cite{ribeiro2020automatic}&0.8747 & 0.9100   & 0.9958 & 0.8872 \\\Xhline{1.5pt}
			\uppercase\expandafter{\romannumeral1} & 0.3971 & 0.1309 & 0.9910 & 0.1824  \\ \hline
			MCMA+\uppercase\expandafter{\romannumeral1} & 0.8289 & 0.8257 & 0.9956 & 0.8223  \\ \hline
			MCMA GAIN & 0.4318 & 0.6948 & 0.0046 & 0.6399  \\ \Xhline{1.5pt}
			\uppercase\expandafter{\romannumeral2} & 0.0682 & 0.0339 & 0.9778 & 0.0333  \\ \hline
			MCMA+\uppercase\expandafter{\romannumeral2} & 0.8401 & 0.8588 & 0.9946 & 0.8410  \\ \hline
			MCMA GAIN & 0.7719 & 0.8249 & 0.0168 & 0.8077  \\ \Xhline{1.5pt}
			\uppercase\expandafter{\romannumeral3} & 0.1667 & 0.0056 & 0.9998 & 0.0108  \\ \hline
			MCMA+\uppercase\expandafter{\romannumeral1} & 0.7237 & 0.6784 & 0.9923 & 0.6840  \\ \hline
			MCMA GAIN & 0.5570 & 0.6728 & -0.0075 & 0.6732  \\ \Xhline{1.5pt}
			aVR & 0.0000 & 0.0000 & 0.9985 & 0.0000  \\ \hline
			MCMA+aVR  & 0.4775 & 0.4261 & 0.9816 & 0.4348  \\ \hline
			MCMA GAIN & 0.4775 & 0.4261 & -0.0169 & 0.4348  \\ \Xhline{1.5pt}
			aVL & 0.0000 & 0.0000 & 0.9998 & 0.0000  \\ \hline
			MCMA+aVR  & 0.5728 & 0.6390 & 0.9827 & 0.5905  \\ \hline
			MCMA GAIN & 0.5728 & 0.6390 & -0.0171 & 0.5905  \\ \Xhline{1.5pt}
			aVF & 0.0000 & 0.0000 & 1.0000 & 0.0000  \\ \hline
			MCMA+aVR  & 0.5226 & 0.6532 & 0.9706 & 0.5223  \\ \hline
			MCMA GAIN & 0.5226 & 0.6532 & -0.0294 & 0.5223  \\ \Xhline{1.5pt}

			V1 & 0.2641 & 0.2510 & 0.9973 & 0.2573  \\ \hline
			MCMA+V1 & 0.7670 & 0.8776 & 0.9923 & 0.8146  \\ \hline
			MCMA GAIN & 0.5029 & 0.6266 & -0.0050 & 0.5573  \\ \Xhline{1.5pt}
			V2 & 0.1667 & 0.0611 & 1.0000 & 0.0894  \\ \hline
			MCMA+V2 & 0.7377 & 0.8435 & 0.9915 & 0.7824  \\ \hline
			MCMA GAIN & 0.5710 & 0.7824 & -0.0085 & 0.6930  \\ \Xhline{1.5pt}
			V3 & 0.2428 & 0.1267 & 0.9990 & 0.1469  \\ \hline
			MCMA+V3 & 0.7669 & 0.8438 & 0.9929 & 0.8006  \\ \hline
			MCMA GAIN & 0.5241 & 0.7171 & -0.0061 & 0.6537  \\ \Xhline{1.5pt}
			V4 & 0.1667 & 0.0090 & 1.0000 & 0.0171  \\ \hline
			MCMA+V4 & 0.7943 & 0.8373 & 0.9936 & 0.8090  \\ \hline
			MCMA GAIN & 0.6276 & 0.8283 & -0.0064 & 0.7919  \\ \Xhline{1.5pt}
			V5 & 0.0000 & 0.0000 & 1.0000 & 0.0000  \\ \hline
			MCMA+V5 & 0.7582 & 0.8285 & 0.9917 & 0.7854  \\ \hline
			MCMA GAIN & 0.7582 & 0.8285 & -0.0083 & 0.7854  \\ \Xhline{1.5pt}
			V6 & 0.0833 & 0.0049 & 0.9996 & 0.0093  \\ \hline
			MCMA+V6 & 0.7450 & 0.8113 & 0.9921 & 0.7717  \\ \hline
			MCMA GAIN & 0.6617 & 0.8064 & -0.0075 & 0.7624  \\ \Xhline{1.5pt}
	\end{tabular}}
	\label{Table:mean_diagnostic}
\end{table}

\subsection{Comparison with Other Methods}
\par MCMA compares with other research works, including Garg  et al. \cite{garg2023single}, Seo  et al. \cite{seo2022multiple}, and Joo  et al. \cite{joo2023twelve}. As known, Garg et al  \cite{garg2023single} adopts the lead \uppercase\expandafter{\romannumeral2}, while Seo  et al. \cite{seo2022multiple} and Joo  et al. \cite{joo2023twelve} utilizes the lead \uppercase\expandafter{\romannumeral1}. Moreover, MCMA can convert arbitrary single-lead ECG into the standard 12-lead ECG. The comparisons in signal-level, feature-level, and diagnostic-level are shown in Table \ref{Table:compare_signal}, Table \ref{Table:compare_feature} and Table \ref{Table:compare_diagnostic}.

\begin{table*}[htbp]
	\centering
	\caption{ {The signal-level comparison of different methods in PTB-XL and CPSC2018}}
	
	\setlength{\tabcolsep}{0.55mm}{
		\tiny
		\begin{tabular}{c|c|ccccccccccccccc}
			\Xhline{1.5pt}
			Dataset&Metric&Method&Input& \uppercase\expandafter{\romannumeral1}& \uppercase\expandafter{\romannumeral2}& \uppercase\expandafter{\romannumeral3}&aVR&aVL&aVF&V1&V2&V3&V4&V5&V6&Mean	
			\\
			\hline
			PTB-XL&{$MSE$}&Garg  et al. \cite{garg2023single}& \uppercase\expandafter{\romannumeral2}& 0.0122 &\redbold{ 0.0001} & \redbold{0.0123} & \redbold{0.0031} & \redbold{0.0122} & \redbold{0.0031} & \redbold{0.0323} & \redbold{0.0690} & {0.0735} & \redbold{0.0477} & \redbold{0.0374} & \redbold{0.0474} & \redbold{0.0292} \\
			&&MCMA&\uppercase\expandafter{\romannumeral2} &\redbold{0.0119} & 0.0093 & 0.0203 & 0.0058 & 0.0138 & 0.0118 & 0.0341 & 0.0691 &\redbold{ 0.0721} & 0.0480 & 0.0388 & 0.0483 & 0.0319 \\         
			&&Seo  et al. \cite{seo2022multiple}& \uppercase\expandafter{\romannumeral1}&	   \redbold{0.0001} & \redbold{0.0148} &\redbold{ 0.0149} & \redbold{0.0037} & \redbold{0.0038} & \redbold{0.0148} &\redbold{ 0.0263} & \redbold{0.0579} & \redbold{0.0640} & \redbold{0.0477} & \redbold{0.0388} & \redbold{0.0482} & \redbold{0.0279} \\
			&&Joo  et al. \cite{joo2023twelve}& \uppercase\expandafter{\romannumeral1}&	  0.0002 & 0.0189 & 0.0187 & 0.0050 & 0.0059 & 0.0190 & 0.0317 & 0.0868 & 0.0920 & 0.0686 & 0.0517 & 0.0546 & 0.0378 \\
			&&Zhan  et al. \cite{zhan2024conditional}& \uppercase\expandafter{\romannumeral1}&0.0003 & 0.0153 & 0.0154 & 0.0041 & 0.0042 & 0.0153 & 0.0267 & 0.0589 & 0.0660 & 0.0495 & 0.0391 & 0.0483 & 0.0286\\
			&&MCMA& \uppercase\expandafter{\romannumeral1} & 0.0071 & 0.0166 & 0.0219 & 0.0064 & 0.0106 & 0.0175 & 0.0326 & 0.0679 & 0.0698 & 0.0482 & 0.0398 & 0.0491 & 0.0323   \\ 
			
			\cmidrule{2-17}
			&{$PCC$}&Garg  et al. \cite{garg2023single}& \uppercase\expandafter{\romannumeral2}&			0.8087 & \redbold{0.9986} & {0.6409} & {0.9330}  & 0.6029 & {0.8873} & 0.7792 & {0.7344} & 0.7056 & 0.7959 & 0.8482 & 0.8427 & 0.7981 \\
			&&MCMA& \uppercase\expandafter{\romannumeral2} & \redbold{ 0.8358} & {0.9897} & \redbold{0.6738} & \redbold{0.9366} & \redbold{0.6519} & \redbold{0.8887} & \redbold{0.8095} & \redbold{0.7833} & \redbold{0.7516} & \redbold{0.8167} & \redbold{0.8599} & \redbold{0.8572} & \redbold{0.8212 } \\ 
			&&Seo  et al. \cite{seo2022multiple}& \uppercase\expandafter{\romannumeral1}&\redbold{0.9983} & 0.7527 & \redbold{0.5849} & \redbold{0.9189} & \redbold{0.8576} & {0.5407} & 0.8237 & {0.7805} & {0.7472} & 0.7931 & 0.8346 & 0.8295 & {0.7885} \\
			&&Joo  et al. \cite{joo2023twelve}& \uppercase\expandafter{\romannumeral1}&0.9962 & 0.6807 & 0.4571 & 0.8905 & 0.8178 & 0.4021 & 0.7827 & 0.6928 & 0.6416 & 0.7112 & 0.7837 & 0.7827 & 0.7199 \\
			&&Zhan  et al. \cite{zhan2024conditional}& \uppercase\expandafter{\romannumeral1}&0.9923 & 0.7332 & 0.5673 & 0.9093 & 0.8390 & 0.5182 & 0.8172 & {0.7811} & 0.7415 & 0.7818 & 0.8290 & 0.8246 & 0.7779  \\
			&&MCMA& \uppercase\expandafter{\romannumeral1} &0.9876 & \redbold{0.7588} & {0.5714} & {0.9167} & {0.8374} & \redbold{0.5516} & \redbold{0.8297} & \redbold{0.7950} & \redbold{0.7646} & \redbold{0.8075} & \redbold{0.8430}& \redbold{0.8358} & \redbold{0.7916}  \\ \hline
			\hline

			\hline
			CPSC2018&{$MSE$}&Garg  et al. \cite{garg2023single}& \uppercase\expandafter{\romannumeral2}&\redbold{0.0215} & \redbold{0.0024} & \redbold{0.0289} & 0.0668 & \redbold{0.0306} & \redbold{0.0101} & \redbold{0.0958}& \redbold{0.1232} & \redbold{0.1532} & \redbold{0.1651} & \redbold{0.2047} & \redbold{0.2579} & \redbold{0.0967} \\
			&&MCMA& \uppercase\expandafter{\romannumeral2}&           0.0369 & 0.0339 & 0.0413 &\redbold{ 0.0312} & 0.0364 & 0.0336 & 0.0938 & 0.1340 & 0.1592 & 0.1716 & 0.2135 & 0.2698 & 0.1046  \\ 
			
			&&Seo  et al. \cite{seo2022multiple}& \uppercase\expandafter{\romannumeral1}&\redbold{0.0036} & \redbold{0.0315} & \redbold{0.0454} & {0.0408} & \redbold{0.0162} & \redbold{0.0339} & {0.0861} &\redbold{ 0.1171} & \redbold{0.1488} & \redbold{0.1694} & \redbold{0.2084} & \redbold{0.2655} &\redbold{ 0.0972} \\
			&&Joo  et al. \cite{joo2023twelve}& \uppercase\expandafter{\romannumeral1}&0.0108 & 0.0409 & 0.0455 & 0.0371 & 0.0228 & 0.0428 & 0.0993 & 0.1834 & 0.1827 & 0.1821 & 0.2185 & 0.2755 & 0.1118 \\
			&&Zhan  et al. \cite{zhan2024conditional}& \uppercase\expandafter{\romannumeral1}&0.0141 & 0.0393 & 0.0443 & \redbold{0.0304} & 0.0237 & 0.0403 & \redbold{0.0838} & 0.1225 & 0.1571 & 0.1759 & 0.2149 & 0.2704 & 0.1014  \\ 
			
			&&MCMA& \uppercase\expandafter{\romannumeral1}&0.0311 & 0.0424 & 0.0447 & 0.0321 & 0.0322 & 0.0406 & 0.0892 & 0.1277 & 0.1544 & 0.1707 & 0.2133 & 0.2700 & 0.1040  \\ 
			\cmidrule{2-17}

			&{$PCC$}&Garg  et al. \cite{garg2023single}& \uppercase\expandafter{\romannumeral2}&0.7552 & \redbold{0.9981} & {0.5686} & 0.9264 & {0.4242} & \redbold{0.8901} & 0.6225 & 0.5803 & 0.6329 & 0.7861 & 0.8376 & 0.8369 & 0.7382 \\
			&&MCMA& \uppercase\expandafter{\romannumeral2}& \redbold{0.7857} & 0.9874 & \redbold{0.5924} & \redbold{0.9347} & \redbold{0.4796} & {0.8900} & \redbold{0.6677} & \redbold{0.6464} & \redbold{0.6868} & \redbold{0.8044} & \redbold{0.8505} & \redbold{0.8521} & \redbold{0.7648}  \\ 
			
			&&Seo  et al. \cite{seo2022multiple}& \uppercase\expandafter{\romannumeral1}&\redbold{0.9978} & 0.7402 & 0.4153 & 0.8932 & \redbold{0.7442} & 0.4830  & 0.7129 & 0.6268 & 0.6619 & 0.7828 & 0.8378 & 0.8382 & 0.7278 \\
			&&Joo  et al. \cite{joo2023twelve}& \uppercase\expandafter{\romannumeral1}&0.9950  & 0.7303 & 0.3517 & 0.8827 & 0.7181 & 0.4701 & 0.6453 & 0.5385 & 0.5238 & 0.7481 & 0.8104 & 0.8002 & 0.6845 \\
			&&Zhan  et al. \cite{zhan2024conditional}& \uppercase\expandafter{\romannumeral1}&0.9889 & 0.7246 & 0.4069 & 0.8845 & 0.7351 & 0.4805 & 0.7071 & 0.6475 & 0.6538 & 0.7636 & 0.8313 & 0.8274 & 0.7209\\
			&&MCMA& \uppercase\expandafter{\romannumeral1}&       0.9823 & \redbold{0.7645} & \redbold{0.4415} & \redbold{0.8997} & {0.7368} & \redbold{0.5368} & \redbold{0.7334} & \redbold{0.7064} & \redbold{0.7289} & \redbold{0.8097} & \redbold{0.8521} & \redbold{0.8515} & \redbold{0.7536}   \\ 
			\Xhline{1.5pt}
	\end{tabular}}
	\vspace{0.5cm}
	\label{Table:compare_signal}
\end{table*}
\begin{table*}[htbp]
	\centering
	\caption{ {The feature-level comparison of different methods in PTB-XL and CPSC2018}}
	
	\setlength{\tabcolsep}{4.8mm}{
		
		\begin{tabular}{cccccc}
			\Xhline{1.5pt}
			Dataset&Method&Input&{$MHR_{SD}$}&{$MHR_{CV}$}&{$MHR_{Range}$}
			\\
			\hline
			PTB-XL&Original&*&2.2137&3.21\%&7.2195\\
			&Garg  et al. \cite{garg2023single}&Lead \uppercase\expandafter{\romannumeral2}&
			{1.1608}&{1.70\%}&{3.5872}\\
			&MCMA&Lead \uppercase\expandafter{\romannumeral2}&\redbold{1.0073}&\redbold{1.51}\%&\redbold{3.0958}\\
			
			&Seo  et al. \cite{seo2022multiple}&Lead \uppercase\expandafter{\romannumeral1}&
			1.8943&2.74\%&6.3984\\
			&Joo  et al. \cite{joo2023twelve}&Lead \uppercase\expandafter{\romannumeral1}&
			2.6891&4.03\%&8.8273\\
			&Zhan  et al. \cite{zhan2024conditional}&Lead \uppercase\expandafter{\romannumeral1}&2.6952&3.82\%&9.0689\\
			&MCMA&Lead \uppercase\expandafter{\romannumeral1}&\redbold{1.1164}&\redbold{1.63\%}&\redbold{3.5413}\\
			
			\hline
			CPSC2018&Original&*&2.1313&2.65\%&7.1267\\ 
			&Garg  et al. \cite{garg2023single}&Lead \uppercase\expandafter{\romannumeral2}&
			{0.9545}&{1.24\%}&{3.0523}\\
			&MCMA&Lead \uppercase\expandafter{\romannumeral2}&\redbold{0.7732}&\redbold{0.99\%}&\redbold{2.4776}\\
			&Seo  et al. \cite{seo2022multiple}&Lead \uppercase\expandafter{\romannumeral1}&
			2.1899&2.79\%&7.5269\\
			&Joo  et al. \cite{joo2023twelve}&Lead \uppercase\expandafter{\romannumeral1}&
			2.4136&3.31\%&8.1059\\
			&Zhan  et al. \cite{zhan2024conditional}&Lead \uppercase\expandafter{\romannumeral1}&2.8610&3.71\%&9.9589\\
			&MCMA& Lead \uppercase\expandafter{\romannumeral1}&\redbold{1.3510}&\redbold{1.40\%}&\redbold{4.4984}\\
			\Xhline{1.5pt}
	\end{tabular}}
	\label{Table:compare_feature}
\end{table*}

\begin{table*}[htbp]
	\centering
	\caption{ {The diagnostic-level comparison of different methods in CODE-test }}
	
	\setlength{\tabcolsep}{5.2mm}{
		
		\begin{tabular}{cccccc}
			\Xhline{1.5pt}
			Method&Input&{$Pre$}&{$Rec$}&{$Spe$}&{$F_{1}$}\\
			\hline
			Reference& 12-lead ECG&0.8747 & 0.9100   & 0.9958 & 0.8872\\ 
			Garg  et al. \cite{garg2023single}&Lead \uppercase\expandafter{\romannumeral2}&0.7268&	\redbold{0.8542}	&0.9881&	0.7808\\
			Input for MCMA&Lead \uppercase\expandafter{\romannumeral2}&0.0682	&0.0339	&0.9778	&0.0333\\
			MCMA&Lead \uppercase\expandafter{\romannumeral2}&\redbold{0.8401}	&0.8588&	\redbold{0.9946}	&\redbold{0.8410}\\
			Seo  et al. \cite{seo2022multiple}&Lead \uppercase\expandafter{\romannumeral1}&0.8248&	0.8480	&0.9948	&0.8299\\
			Joo  et al. \cite{joo2023twelve}&Lead \uppercase\expandafter{\romannumeral1}&0.7817&	0.7846&	0.9938&	0.7730\\
			Zhan et al  \cite{zhan2024conditional}&Lead \uppercase\expandafter{\romannumeral1}&0.8171&\redbold{0.8739}&0.9946&\redbold{0.8423}\\
			Input for MCMA&Lead \uppercase\expandafter{\romannumeral1}&0.3971&	0.1309	&0.9910	&0.1824\\
			MCMA&Lead \uppercase\expandafter{\romannumeral1}&\redbold{0.8289}&	0.8257	&\redbold{0.9956}	&{0.8223}\\
			
			\Xhline{1.5pt}
	\end{tabular}}
	\label{Table:compare_diagnostic}
\end{table*}

\subsection{Ablation Study}
\par MCMA utilizes two key modules, one for arbitrary single-lead ECG reconstruction, and another for zero-padding strategy. Then, it is necessary to compare with different settings, including fixed-channel(lead \uppercase\expandafter{\romannumeral1} as an example) and copy-padding strategy. The signal-level evaluation metric includes mean square error ({$MSE$}) and Pearson correlation coefficient ({$PCC$}). The experimental results comparison with different settings can be shown in Table \ref{Table:MCMA}, including the lead \uppercase\expandafter{\romannumeral1} and the average value for 12 single-lead ECG. In the most cases, MCMA has achieved excellent result in 12-lead ECG reconstruction task.

\begin{table*}[htbp]
	\centering
	\caption{ {The ablation study for the proposed framework, MCMA, which adopts the zero-padding strategy and supports arbitrary single-lead ECG as input}}
	\setlength{\tabcolsep}{5.mm}{
		\begin{tabular}{c|cccccc}
			\Xhline{1.5pt}
			\multicolumn{3}{c}{Setting}&\multicolumn{2}{c}{PTB-XL}&\multicolumn{2}{c}{CPSC2018}\\\hline
			Arbitrary&Padding&Input& {$MSE$}& {$PCC$}& {$MSE$}& {$PCC$}\\\hline
			
			No&Zeros&Lead \uppercase\expandafter{\romannumeral1} &\redbold{0.0322}&0.7896&0.1043&\redbold{0.7541}\\
			Yes&Copy&Lead \uppercase\expandafter{\romannumeral1}&0.0326&0.7742&0.1055&0.7058\\
			Yes&Zeros&Lead \uppercase\expandafter{\romannumeral1} &{0.0323}&\redbold{0.7916}&\redbold{0.1040}&{0.7536}\\
			\hline
			No&Zeros&12 Single-lead&0.0575&0.3402&0.1318&0.3053\\
			Yes&Copy&12 Single-lead&0.0329&0.7602&0.1068&0.6569\\
			Yes&Zeros&12 Single-lead &\redbold{0.0317}&\redbold{0.7885}&\redbold{0.1034}&\redbold{0.7420}\\
			\hline
			\Xhline{1.5pt}
	\end{tabular}}
	\label{Table:MCMA}
\end{table*}
\subsection{Case Study}
\par
\begin{figure}[htbp]
	\centering
	\includegraphics[width=0.45\textwidth]{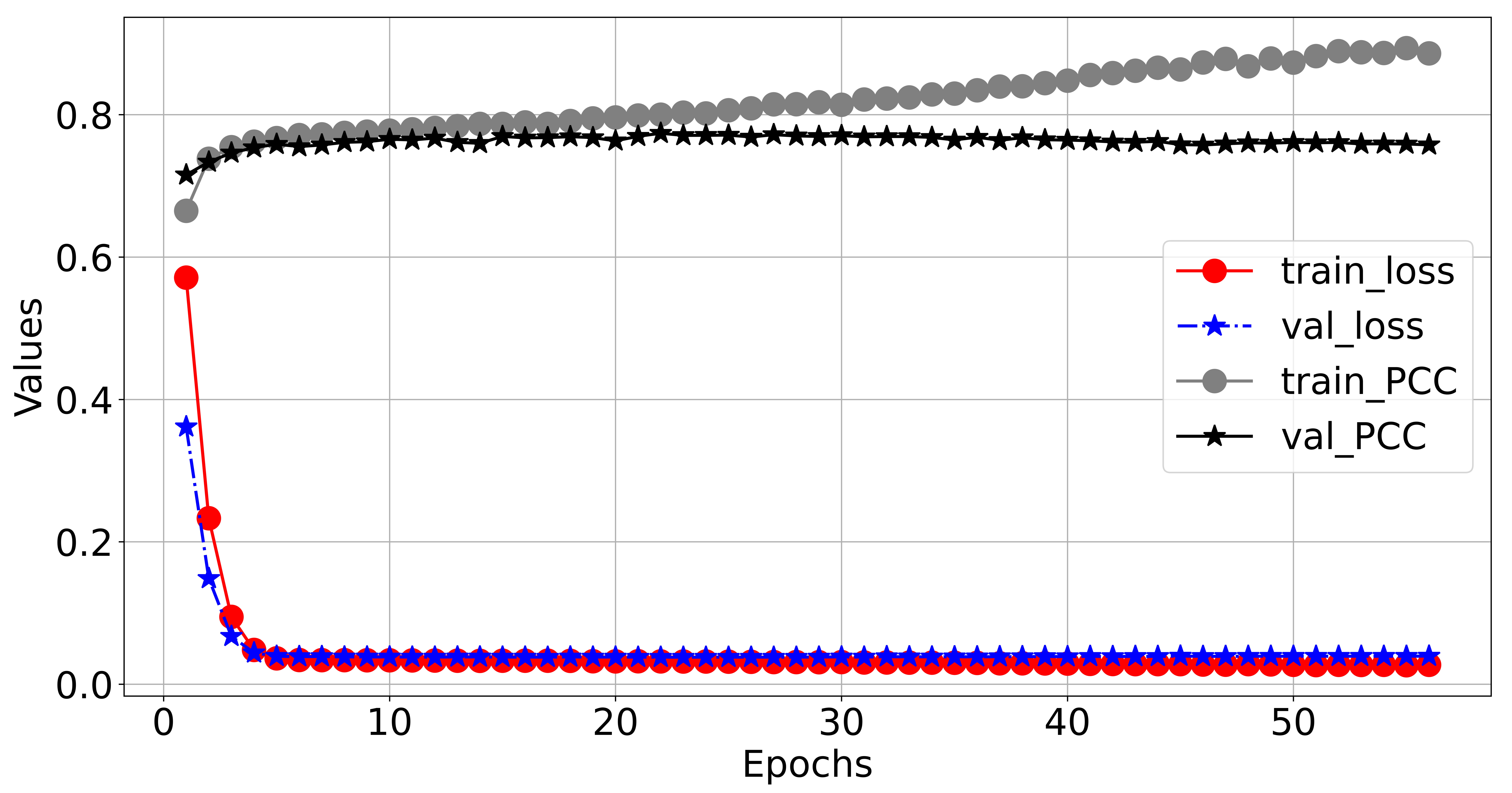}
	\caption{The mean square error and Pearson correlation 
		coefficient in the training process. The red circle means training loss, the blue star is validation loss, the black circle means training Pearson correlation 
		coefficient ({$PCC$}), and the black star means validation Pearson correlation 
		coefficient ({$PCC$}).}
	\label{fig:training}
\end{figure}

\begin{figure*}[htbp]
	\centering
	\includegraphics[width=0.95\textwidth]{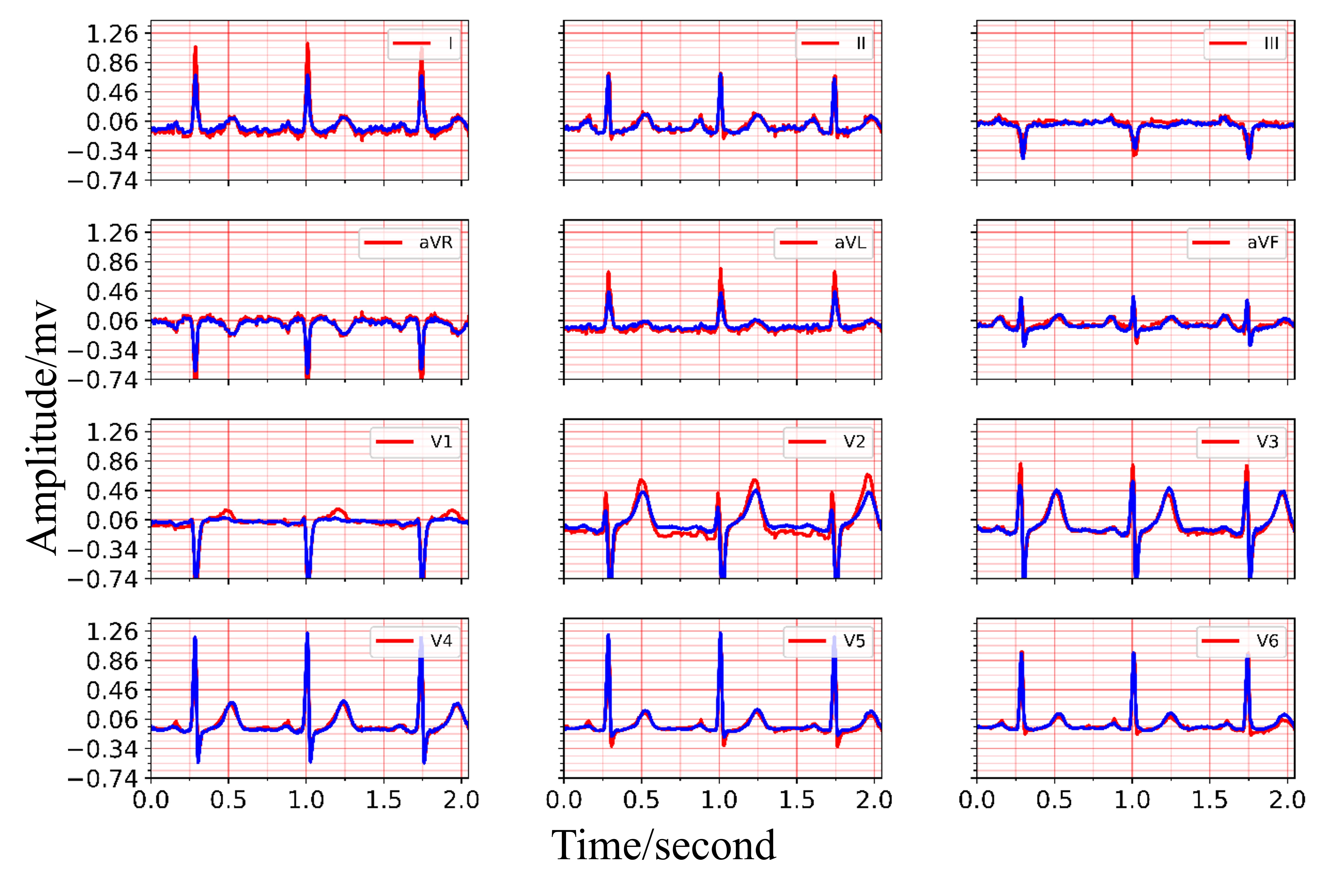}
	\caption{The 12-lead ECG reconstruction performance in the internal testing set PTB-XL, the red lines are the real signals while the blue lines represent the generated signals.}
	\label{fig:signal_ptbxl}
	\vspace{-0.25cm}
\end{figure*}

\begin{figure*}[htbp]
	\centering
	\includegraphics[width=0.95\textwidth]{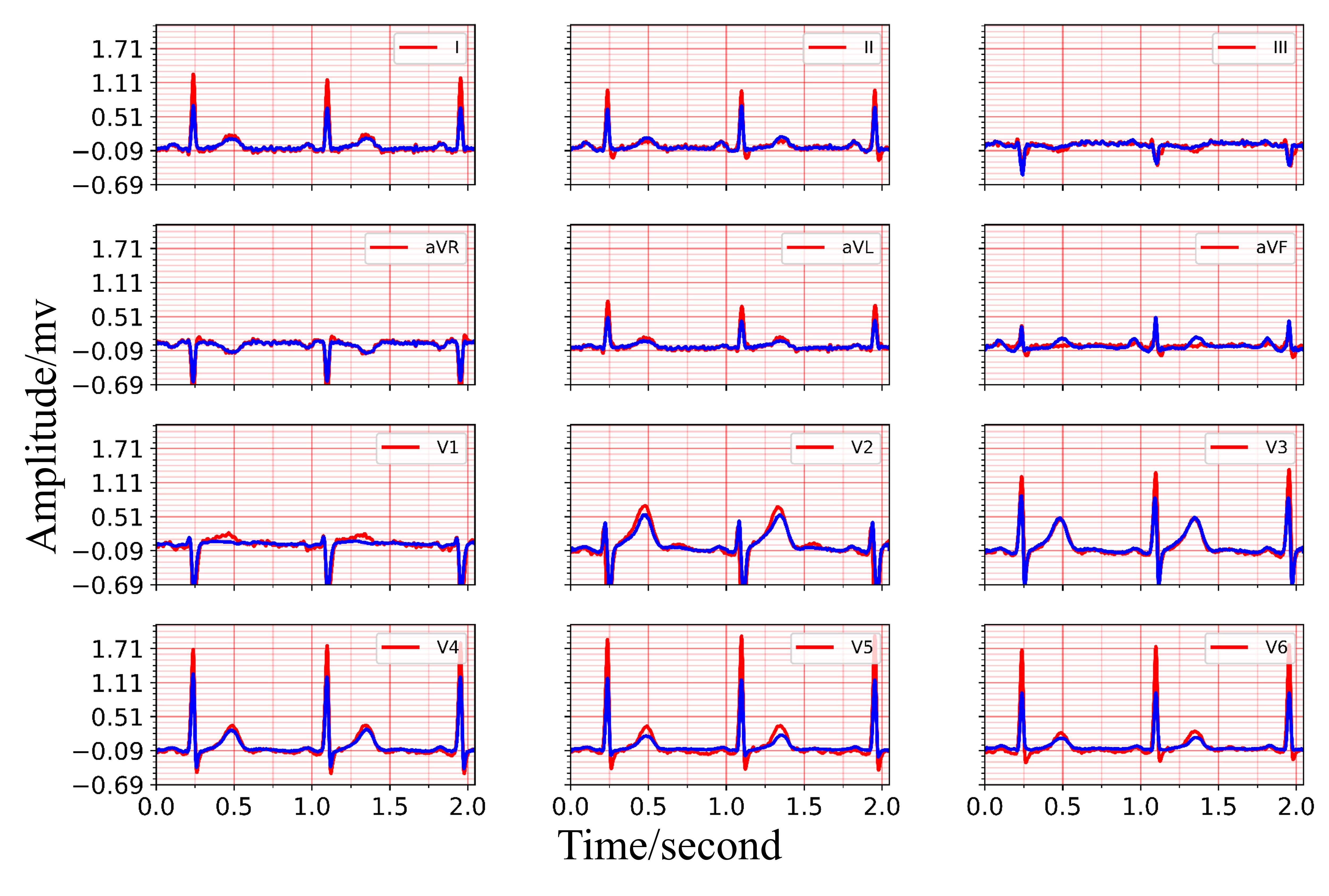}
	\caption{The 12-lead ECG reconstruction performance in the external testing set CPSC2018, the red lines are the real signals while the blue lines represent the generated signals.}
	\label{fig:signal_cpsc}
	\vspace{-0.25cm}
\end{figure*}

 The training process details of MCMA can be illustrated as seen in Fig.\ref{fig:training}. To show the advantages of the proposed framework, the generated and real 12-lead ECG should be clearly shown in Fig.\ref{fig:signal_ptbxl}, in which the generated and the real signals are colored blue and red. Fig.\ref{fig:signal_ptbxl} demonstrates the great generation ability of the proposed framework. For example, the average {$MSE$} and {$PCC$} between the generated and real 12-lead ECG is 0.0032 and 0.9560, and it is concluded that the generator can generate 12-lead ECG with single-lead ECG. Besides the internal testing dataset (i.e., PTB-XL), the external testing dataset's (i.e., CPSC2018) reconstruction performance demonstrates the proposed framework's advantages from another aspect, as seen in Fig.\ref{fig:signal_cpsc}.
\begin{figure*}[htbp]
	\centering
	\includegraphics[width=0.925\textwidth]{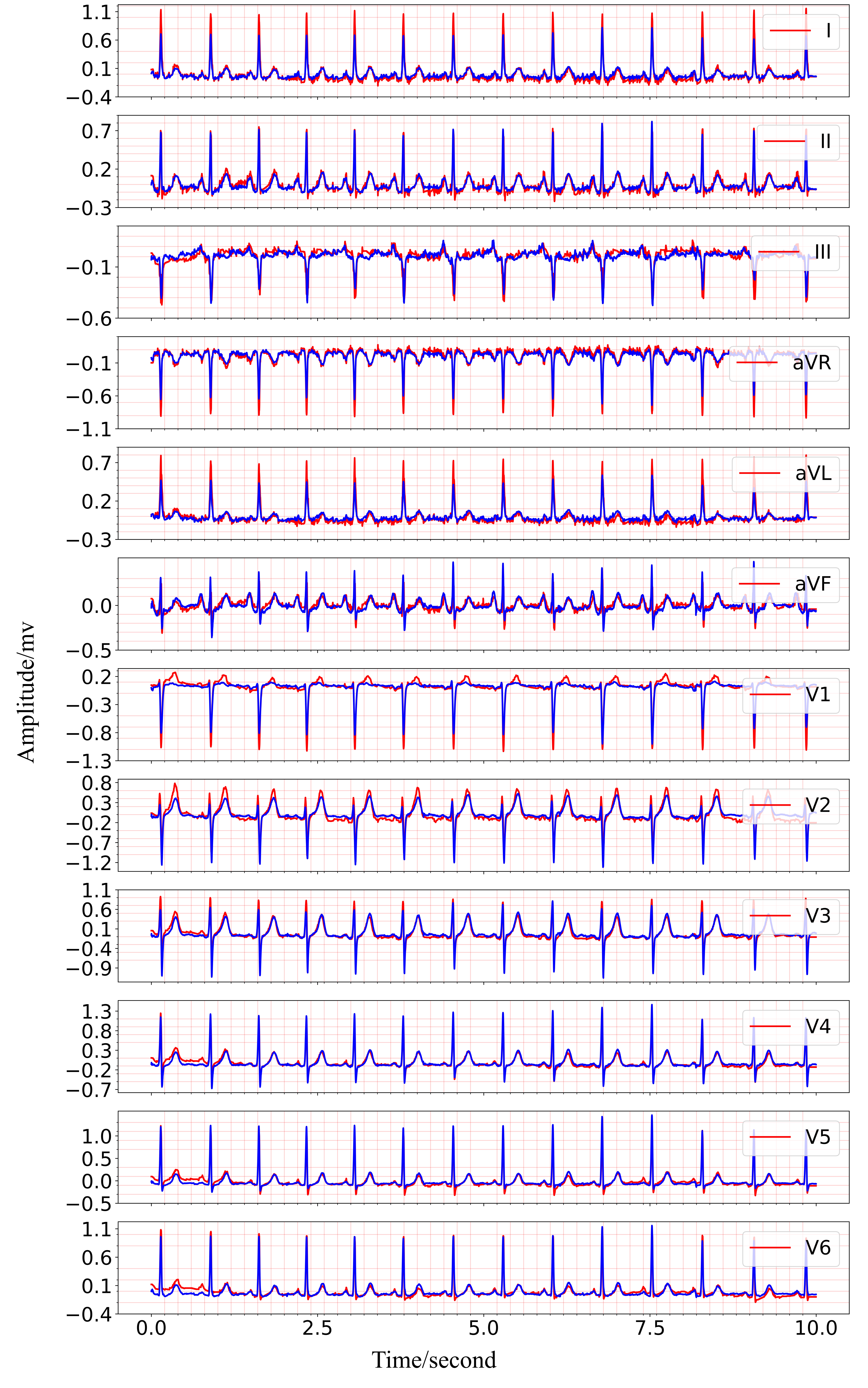}
	\caption{The generated and real 10-second 12-lead ECG, demonstrating its advantages for variable-duration ECG reconstruction, the red lines are the real signals while the blue lines represent the generated signals.}
	\label{fig:signal_10s}
\end{figure*}
\par Based on the experimental result provided in Fig.\ref{fig:signal_ptbxl} and Fig.\ref{fig:signal_cpsc}, it is shown that the multi-channel masked autoencoder (MCMA) can be used to reconstruct the 12-lead ECG with single-lead ECG. In the clinical practice, the ECG collected by wearable devices can be with different signal length, instead of the fixed length. It is necessary to demonstrate the proposed framework cloud also works with the variable-duration ECG signals, and the signal reconstruction result with 10-second ECG is seen in Fig.\ref{fig:signal_10s}. In this case, the 5000 points should be filled with the extra 120 points, and it can be as the 5 individual samples for MCMA to reconstruct 12-lead ECG with single-lead ECG as input.

\section{Discussion}

\par In this work, we propose a multi-channel masked autoencoder, MCMA, for generating the standard 12-lead ECG with arbitrary single-lead ECG. Further, this study establishes a comprehensive evaluation benchmark, ECGGenEval, including the signal-level, feature-level, and diagnostic-level evaluation.
\par MCMA can work well in ECGGenEval, achieving the state-of-the-art performance. MCMA can convert arbitrary single-lead ECG into 12-lead ECG, instead of the fixed-lead ECG \cite{garg2023single,joo2023twelve,seo2022multiple,zhan2024conditional}. Secondly, we provide multiple level evaluation results in an internal and two external testing datasets, and the details are as follows.

\par Firstly, according to the signal-level evaluation results from Table \ref{Table:ptbxl_mse_pcc} and Table \ref{Table:cpsc_mse_pcc}, on the mentioned experimental results, it is known that the proposed framework can reconstruct high-fidelity 12-lead ECG with single-lead ECG. The average {$MSE$} and {$PCC$} in PTB-XL are 0.0317 and 0.7885, while the average {$MSE$} and {$PCC$} in CPSC2018 are 0.1034 and 0.7420, respectively. The reconstruction performance in the internal and external testing dataset can demonstrate its advantages, and MCMA can reconstruct the standard 12-lead ECG with arbitrary single-lead ECG as input. Therefore, the proposed method can provide a feasible solution when collecting the standard 12-lead ECG is inconvenient and difficult, like remote cardiac healthcare. In the signal-level comparison, the {$MSE$} and {$PCC$} for generating 12-lead ECG with lead \uppercase\expandafter{\romannumeral2} are 0.0319 and 0.8212, better than Garg et al. \cite{garg2023single}, with the {$MSE$} of 0.0292 and {$PCC$} of 0.7981. At the same time, in the CPSC2018, MCMA can achieve a better {$PCC$} compared with other works. Therefore, MCMA can be used for 12-lead ECG reconstruction tasks while the single-lead ECG is collected, and the signal-level evaluation provides a novel solution in real-world cardiac healthcare applications. 

\par  { Secondly, Table \ref{Table:ptbxl_mhr} and Table \ref{Table:cpsc_mhr} complete the feature-level evaluation. For the internal testing dataset, PTB-XL, Table \ref{Table:ptbxl_mhr} demonstrates that the heart rate estimation in different leads is similar in the generated 12-lead ECG, and it is even better than the original 12-lead ECG. The estimated heart rate from the real 12-lead ECG may be different, since the noise exists in special channels. Table \ref{Table:ptbxl_mhr} shows that the average {$MHR_{SD}$}, {$MHR_{CV}$} and {$MHR_{Range}$} are 1.0481, 1.58\% and 3.2874, in which  the optimal result is from the generated 12-lead ECG by lead V4 ECG. Table \ref{Table:cpsc_mhr} shows the external evaluation in CPSC2018, the average {$MHR_{SD}$}, {$MHR_{CV}$} and {$MHR_{Range}$} are 0.9483, 1.24\% and 3.0916, while the optimal result is from the generated 12-lead ECG by lead \uppercase\expandafter{\romannumeral2} ECG. The generated 12-lead ECG from arbitrary single-lead ECG can produce a good heart rate consistency in different leads, and it can even be better than the original 12-lead ECG in some cases, due to the ECG signal denoising function in the proposed framework. Table \ref{Table:compare_feature} demonstrates the advantages of MCMA over others, which can be highlighted as red. Therefore, the feature-level evaluation can demonstrate the advantages of MCMA.} 
\par  {Based on Table \ref{Table:classification_12_1}, the classifier can adopt the generated 12-lead ECG for arrhythmia classification. The average F1-score over 6 classes is 0.8319. Then, it is proven that MCMA can convert the single-lead ECG into the 12-lead ECG, and the generated 12-lead ECG can retain the pathological information, and it is different to the signal-level and feature-level evaluation. Therefore, with the multi-channel masked autoencoder, it is possible to complete arrhythmia classification with single-lead ECG, like lead \uppercase\expandafter{\romannumeral1} ECG in Table \ref{Table:classification_12_1}. Further, according to Table \ref{Table:mean_diagnostic}, the classification performance of the generated 12-lead ECG is better than that of single-lead ECG and similar to the real 12-lead ECG, which can demonstrate the classification performance gain brought by MCMA. The generated 12-lead from lead \uppercase\expandafter{\romannumeral1} can provide the closest classification performance, the average {$F_{1}$} is 0.8319, which exceeds other cases. According to Table \ref{Table:compare_diagnostic}, the classification performance with generated 12-lead ECG is improved. For example, taking lead \uppercase\expandafter{\romannumeral2} as input, Garg  et al. \cite{garg2023single}can achieve a {$F_{1}$} of 0.7807, lower than the proposed method. Similarly, with the lead \uppercase\expandafter{\romannumeral1} as input, Seo  et al. \cite{seo2022multiple} and Joo  et al. \cite{joo2023twelve} have a {$F_{1}$} of 0.8299 and 0.7730, respectively, while MCMA can be with a {$F_{1}$} with 0.8223. From the view of classification task, the classification performance in the above tables demonstrates the generated 12-lead ECG can be used for cardiac abnormality detection, which can prove its advantage in bridging the single-lead ECG and 12-lead ECG, and it is effective to generate the pathological information with single-lead ECG as input. } 

\par  { As Table \ref{Table:MCMA} showing, the proposed framework is effective. The multi-channel strategy can support arbitrary single-lead to generate 12-lead ECG. Although the reconstruction performance of lead \uppercase\expandafter{\romannumeral1} is slightly lower than the fixed-channel. When the lead \uppercase\expandafter{\romannumeral1} ECG inputs, the fixed-channel can have a {$MSE$} of 0.0322 and a {$PCC$} of 0.7896, while MCMA can be with a {$MSE$} of 0.0323 and a {$PCC$} of 0.7916. However, for the fixed-channel, it is difficult to realize 12-lead ECG reconstruction with other leads, and the training and inference cost is largely different in training and storing 12 models with this setting. Further, the zero-padding strategy is better than the copy-padding strategy, while the two strategies both support the 12-lead reconstruction with arbitrary single-lead ECG. The mean {$MSE$} and {$PCC$} in MCMA are 0.0317 and 0.7885, while the mean {$MSE$} and {$PCC$} in copy-padding are 0.0329 and 0.7602, respectively}. 
\par  {This study is with the following advantages, from the engineering and clinical perspectives. Firstly, the generated signal is similar to the original signal, as the mean square errors of 0.0317 and 0.1034, correlation coefficients of 0.7885 and 0.7420 in the signal-level evaluation. Secondly, the generated signal can be used in the arrhythmia classification, as the average {$F_{1}$} with two generated 12-lead ECG is 0.8233 and 0.8410 in the diagnostic-level evaluation. According to the mentioned advantages, the contributions are as follows:

\par  Further, this study is expected to be a feasible solution for wearable ECG monitoring, and it is able to improve the clinical importance of arbitrary single-lead ECG. For this research project, these experimented is conducted in these public datasets, such as PTB-XL and CPSC2018. Naturally, there are some limitations in this study, and these issues should be addressed in the future, as follows. High-quality electrocardiogram (ECG) signal acquisition method can significantly impact the reconstruction performance, and it may be addressed in the sensing layer \cite{zhang2024three} or the algorithmic layer \cite{DDPM}. The generated signals necessitate evaluation by professional clinicians to ascertain their viability as a long-term substitute for the conventional 12-lead ECG in continuous monitoring scenarios.In other words, the question is whether a physician can render an equivalent diagnosis utilizing the 12-lead ECG generated by MCMA. Consequently, additional research endeavors are essential to advance the mentioned problems, ultimately realizing the considerable clinical relevance and practical utility.
\section{Conclusion}
\par In a word, this study proposes a novel generative framework to reconstruct 12-lead ECG with a single-lead ECG, as multi-channel masked autoencoder (MCMA), and it involves two main contributions. Firstly, unlike other methods, the proposed framework can convert arbitrary single-lead ECG into the standard 12-lead ECG. The experimental results showed that the proposed framework had excellent performance, achieving state-of-the-art performance on the proposed benchmark, ECGGenEval, including the signal-level, feature-level, and diagnostic-level evaluation.  {For example, the average Pearson correlation coefficients in the internal and external testing set are 0.7885 and 0.7420, outperformed the related approaches. Additionally, it is shown that the zero-padding strategy can play an important role in the proposed framework, beats the copy-padding strategy. In the future, it is necessary to study high-quality ECG and clinical validation, to let the proposed framework play an important role in clinical practice, which provides a novel feasible solution for long-term cardiac health monitoring. }

\section*{Data Availability}
\par All datasets used in this study are openly available. PTB-XL: \url{https://physionet.org/content/ptb-xl/1.0.3/}, CPSC-2018: \url{http://2018.icbeb.org/Challenge.html}, CODE-test: \url{https://zenodo.org/records/3765780}.

\section*{Code Availability}
\par The open-source code is publicly available at \url{https://github.com/CHENJIAR3/MCMA}.
\section*{Acknowledgements}
\par This work was supported by National Natural Science Foundation of China (No. 62102008); Clinical Medicine Plus X - Young Scholars Project of Peking University, the Fundamental Research Funds for the Central Universities (PKU2024LCXQ030). 
\section*{Author Contribution}
\par J.C. contributed to the formal analysis, methodology, visualization, and original draft writing. W.W. provided supervision for the study, and T.L. also offered supervision. S.H. was responsible for the investigation, supervision, funding acquisition, and resource management.
\section*{Competing Interests}
\par The authors declare no competing interests.

\bibliography{ref}

\newpage
\section*{Appendix}

\par This study do not focus on the hyperparameter searching, but they may have a bit of influence on the 12-lead ECG reconstruction, like kernel size and window size, as seen in Supplementary Table \ref{Table:diff_ks} and Supplementary Table \ref{Table:diff_ws}. Since the proposed contributions could be proven with the value setting, the experimental results are provided as a reference for the related works.

\begin{table}[htbp]
	\centering
	\caption{The reconstruction performance with various kernel size({$k$}), and the used kernel size is 5}
	
	\setlength{\tabcolsep}{1.25mm}{
		\begin{tabular}{ccccccc}
			\Xhline{1.5pt}
			\multirow{2}*{$k$}&\multirow{2}*{FLOPs}&\multirow{2}*{Params}&\multicolumn{2}{c}{PTB-XL}&\multicolumn{2}{c}{CPSC2018}\\
			\cmidrule(r){4-7}
			&&& MSE& PCC& MSE& PCC\\\hline
			1 & 172.4M & 1.4120M &0.5689  & 0.0506& 0.5142 & 0.1240  \\ 
			3 & 502.3M & 4.2078M & 0.7859 & 0.0317&  0.7364&0.1039 \\ 
			5 & 832.1M & 7.0036M &0.7885  &0.0317 &0.7420  &0.1034 \\ 
			7 & 1162M & 9.7994M &  0.7888&0.0316 & 0.7371 & 0.1038\\ 
			9 & 1492M & 12.5952M &  0.7876& 0.0316&  0.7337& 0.1039 \\ 
			11 & 1822M & 15.3910M &  0.7892& 0.0315&  0.7418&0.1036 \\ 
			13 & 2151M & 18.1868M &  0.7768&0.0322 & 0.7365 &0.1040 \\ 
			15 & 2481M & 20.9826M & 0.7876 & 0.0316&  0.7395& 0.1038\\ 
			17 & 2811M & 23.7784M &0.7826  &0.0317 &  0.7305&0.1038  \\ 
			19 & 3141M & 26.5743M &0.7858  &0.0317 & 0.7330 &0.1039 \\ 
			\hline
			\Xhline{1.5pt}
	\end{tabular}}
	
	\label{Table:diff_ks}
\end{table}
\begin{table}[htbp]
	\centering
	\caption{{ The reconstruction performance with various window size({$s$}), and the used window size is 2}}
	\setlength{\tabcolsep}{1.20mm}{
		\begin{tabular}{ccccccc}
			\Xhline{1.5pt}
			\multirow{2}*{$s$}&\multirow{2}*{FLOPs}&\multirow{2}*{Params}&\multicolumn{2}{c}{PTB-XL}&\multicolumn{2}{c}{CPSC2018}\\
			\cmidrule(r){4-7}
			&&& MSE& PCC& MSE& PCC\\\hline
			1 &14371M&7.0036M&  0.7785&0.0321 & 0.7234 &0.1042 \\ 
			2 &832.1M&7.0036M&  0.7885& 0.0317& 0.7420& 0.1034\\ 
			4 &106.7M&7.0036M&  0.7774& 0.0321& 0.7270&0.1043  \\ 
			\hline
			\Xhline{1.5pt}
	\end{tabular}}
	\label{Table:diff_ws}
\end{table}
\par Therefore, based on the above experimental results from Supplementary Table \ref{Table:diff_ks} and Supplementary Table \ref{Table:diff_ws}, it is reasonable and acceptable for this study to take 5 and 2 as the kernel size and window size, respectively. The reconstruction performance with this setting could outperform other setting. Further, this study also release the lead index classification accuracy, if users do not get the lead information, the experimental results can be seen in Supplementary Table \ref{Table:ptbxl_lead_class} and Supplementary Table \ref{Table:cpsc2018_lead_class}.
\begin{table*}[htbp]
	\centering
	\caption{{ The lead classification accuracy in the internal testing dataset}}
	\setlength{\tabcolsep}{0.50mm}{
		\begin{tabular}{c|cccccccccccc}
			\Xhline{1.5pt}
			
			\diagbox{Label}{Predict}& \uppercase\expandafter{\romannumeral1}& \uppercase\expandafter{\romannumeral2}& \uppercase\expandafter{\romannumeral3}&aVR&aVL&aVF&V1&V2&V3&V4&V5&V6	\\ \hline
			\uppercase\expandafter{\romannumeral1} & 10798 & 6 & 0 & 0 & 173 & 0 & 0 & 0 & 0 & 5 & 22 & 11  \\ \hline
			\uppercase\expandafter{\romannumeral2} & 10 & 10833 & 16 & 0 & 0 & 93 & 1 & 2 & 2 & 8 & 26 & 24  \\ \hline
			\uppercase\expandafter{\romannumeral3} & 0 & 15 & 10783 & 1 & 0 & 207 & 0 & 0 & 0 & 0 & 5 & 4  \\ \hline
			aVR & 0 & 0 & 0 & 10998 & 2 & 0 & 14 & 1 & 0 & 0 & 0 & 0  \\ \hline
			aVL & 179 & 0 & 0 & 5 & 10769 & 0 & 11 & 36 & 11 & 1 & 0 & 3  \\ \hline
			aVF & 0 & 317 & 111 & 1 & 0 & 10560 & 3 & 0 & 1 & 4 & 10 & 8  \\ \hline
			V1 & 0 & 0 & 8 & 17 & 0 & 0 & 10809 & 159 & 16 & 5 & 1 & 0  \\ \hline
			V2 & 0 & 0 & 6 & 0 & 0 & 0 & 36 & 10835 & 121 & 13 & 2 & 2  \\ \hline
			V3 & 1 & 0 & 0 & 0 & 0 & 1 & 16 & 136 & 10760 & 82 & 15 & 4  \\ \hline
			V4 & 2 & 5 & 0 & 0 & 0 & 8 & 2 & 6 & 98 & 10628 & 217 & 49  \\ \hline
			V5 & 11 & 12 & 3 & 0 & 6 & 1 & 8 & 6 & 22 & 223 & 10431 & 292  \\ \hline
			V6 & 7 & 19 & 2 & 0 & 7 & 3 & 3 & 5 & 1 & 33 & 356 & 10579  \\ \hline
			\hline
			\Xhline{1.5pt}
	\end{tabular}}
	\label{Table:ptbxl_lead_class}
\end{table*}
\begin{table*}[htbp]
	\centering
	\caption{{ The lead classification accuracy in the external testing dataset}}
	\setlength{\tabcolsep}{0.50mm}{
		\begin{tabular}{c|cccccccccccc}
			\Xhline{1.5pt}
			
			\diagbox{Label}{Predict}& \uppercase\expandafter{\romannumeral1}& \uppercase\expandafter{\romannumeral2}& \uppercase\expandafter{\romannumeral3}&aVR&aVL&aVF&V1&V2&V3&V4&V5&V6	\\ \hline
			\uppercase\expandafter{\romannumeral1}& 54201 & 221 & 4 & 31 & 563 & 19 & 0 & 42 & 37 & 127 & 325 & 257  \\ \hline
			\uppercase\expandafter{\romannumeral2} & 103 & 54352 & 106 & 14 & 2 & 903 & 1 & 5 & 15 & 51 & 138 & 137  \\ \hline
			\uppercase\expandafter{\romannumeral3} & 7 & 428 & 53234 & 17 & 0 & 1960 & 29 & 3 & 14 & 33 & 58 & 44  \\ \hline
			aVR & 15 & 159 & 23 & 55361 & 18 & 15 & 136 & 27 & 5 & 17 & 48 & 3  \\ \hline
			aVL & 853 & 136 & 3 & 124 & 54284 & 13 & 79 & 114 & 70 & 48 & 84 & 19  \\ \hline
			aVF & 10 & 2869 & 561 & 14 & 2 & 52172 & 12 & 4 & 15 & 50 & 72 & 46  \\ \hline
			V1 & 20 & 100 & 10 & 100 & 8 & 7 & 54382 & 978 & 134 & 28 & 45 & 15  \\ \hline
			V2 & 18 & 57 & 1 & 49 & 6 & 0 & 131 & 54394 & 1044 & 85 & 35 & 7  \\ \hline
			V3 & 35 & 44 & 2 & 24 & 3 & 3 & 55 & 425 & 54237 & 845 & 118 & 36  \\ \hline
			V4 & 62 & 41 & 2 & 25 & 1 & 7 & 16 & 49 & 705 & 52865 & 1813 & 242  \\ \hline
			V5 & 61 & 62 & 18 & 18 & 6 & 19 & 40 & 63 & 155 & 1333 & 52715 & 1337  \\ \hline
			V6 & 64 & 62 & 8 & 5 & 8 & 13 & 27 & 17 & 39 & 243 & 2206 & 53135  \\ \hline
			\hline
			\Xhline{1.5pt}
	\end{tabular}}
	\label{Table:cpsc2018_lead_class}
\end{table*}
\par Based on the experimental results, the classification accuracy is 0.9743 in the internal testing dataset, and 0.9633 in the external testing dataset. Therefore, it is possible for MCMA to distinguish and classify the different ECG lead, and it will supplement the missing lead index. 

\end{document}